\documentclass[preprint]{elsarticle}
\makeatletter
\def\ps@pprintTitle{%
	\let\@oddhead\@empty
	\let\@evenhead\@empty
	\def\@oddfoot{\centerline{\thepage}}%
	\let\@evenfoot\@oddfoot}
\makeatother

\usepackage{graphicx}

\usepackage{ amsmath , amssymb , amsthm }

\usepackage[hidelinks]{hyperref}

\usepackage{optidef}

\usepackage{amsfonts}
\usepackage{caption}
\usepackage{comment}
\usepackage{mathrsfs}

\usepackage[dvipsnames]{xcolor}

\biboptions{authoryear}

\usepackage{url} 

\usepackage{graphicx}
\usepackage{blindtext}

\usepackage{pdfpages}

\usepackage[utf8]{inputenc}

\usepackage{amsthm}

\usepackage{algorithm}
\usepackage{algpseudocode}

\usepackage{mathtools}
\mathtoolsset{showonlyrefs=true}

\usepackage{subcaption}
\usepackage{tikz}
\usepackage{graphicx}	

\usepackage{float}

\usepackage{pgfplots}
\definecolor{juliacolor}{RGB}{0,155,250}

\usepackage[section]{placeins}

\usepackage{tabularx}
\usepackage{csvsimple}
\usepackage[textheight=9in,
textwidth=6.5in,
top=1in,
headheight=14pt,
headsep=25pt,
footskip=30pt]{geometry} 

\usepackage{physics}

\usepackage{makecell}

\usepackage{booktabs}

\usepackage{appendix}

\DeclareRobustCommand{\P}[0]{\text{P}}
\DeclareMathAlphabet{\mathscrbf}{OMS}{mdugm}{b}{n}

\DeclareMathOperator{\mytrace}{Tr}

\DeclareMathOperator{\E}{\mathbb{E}}

\usepackage{subcaption}
\usepackage{tikz}
\usepackage{graphicx}

\usepackage{xspace}
\newcommand{\arnold}{Constr-DRKM\xspace}

\usepackage{stackengine}

\begin{document}
	
\begin{frontmatter}

\title{Unsupervised learning of disentangled representations in deep restricted kernel machines with orthogonality constraints}

\author[mymainaddress]{Francesco Tonin\corref{mycorrespondingauthor}}
\cortext[mycorrespondingauthor]{Corresponding author}
\ead{francesco.tonin@esat.kuleuven.be}

\author[mymainaddress]{Panagiotis Patrinos}
\ead{panos.patrinos@esat.kuleuven.be}

\author[mymainaddress]{Johan A. K. Suykens}
\ead{johan.suykens@esat.kuleuven.be}
	
\address[mymainaddress]{Department of Electrical Engineering, ESAT-STADIUS,\\
KU Leuven. Kasteelpark Arenberg 10, B-3001 Leuven, Belgium\\}

\begin{abstract}
We introduce \arnold, a deep kernel method for the unsupervised learning of disentangled data representations. We propose augmenting the original deep restricted kernel machine formulation for kernel PCA by orthogonality constraints on the latent variables to promote disentanglement and to make it possible to carry out optimization without first defining a stabilized objective. After illustrating an end-to-end training procedure based on a quadratic penalty optimization algorithm with warm start, we quantitatively evaluate the proposed method's effectiveness in disentangled feature learning. We demonstrate on four benchmark datasets that this approach performs similarly overall to $\beta$-VAE on a number of disentanglement metrics when few training points are available, while being less sensitive to randomness and hyperparameter selection than $\beta$-VAE. We also present a deterministic initialization of \arnold's training algorithm that significantly improves the reproducibility of the results. Finally, we empirically evaluate and discuss the role of the number of layers in the proposed methodology, examining the influence of each principal component in every layer and showing that components in lower layers act as local feature detectors capturing the broad trends of the data distribution, while components in deeper layers use the representation learned by previous layers and more accurately reproduce higher-level features.
\end{abstract}

\begin{keyword}
	Kernel methods \sep Unsupervised learning \sep Manifold learning \sep Learning disentangled representations
	
\end{keyword}

\end{frontmatter}

\section{Introduction}
\label{sec:intro}

The choice of the features on which a machine learning method is trained on is crucial to achieve good performance. Finding representations of the data that make it easier to train classifiers or other predictors is the goal of representation learning \citep{bengio-rl}. \\

One desirable characteristic of good representations is disentanglement, which means that the learned representation separates the factors of variations in the data \citep{bengio2009, bengio-rl}. In fact, in representation learning it is common to assume a low dimensional multivariate random variable $z$ representing the meaningful factors of variations: a high dimensional observation $x$ is then sampled from the conditional distribution $P(x|z)$. For example, a model trained on faces may learn latent generative factors such as hair style, emotion or the presence of glasses. In this respect, the empirical success of deep learning in supervised learning tasks is often linked to the ability of deep networks to learn meaningful intermediate representations \citep{vincent2010, zeiler2014}.\\

Regarding the unsupervised learning setting, many benefits of employing disentangled representations have been highlighted. For example, they could be useful in i) transfer learning, by reusing meaningful representations on new tasks \citep{bengio-rl}, ii) semi-supervised learning \citep{scholkopf2012,ranzato2007}, iii) few-shot learning \citep{ridgeway2018}, iv) explainability, for instance in the medical domain \citep{Holzinger2019, sarhan2019}, and v) reinforcement learning \citep{lake2017}.\\

However, unsupervised learning of disentangled representations is still a key challenge in artificial intelligence research \citep{bengio-rl,lecun2015a,lake2017}. State-of-the-art models are mostly based on the generative adversarial network (GAN) framework, such as InfoGAN \citep{infogan}, or on variational autoencoders (VAEs) \citep{vae}, including $\beta$-VAEs \citep{betavae}, FactorVAE \citep{factorvae} and $\beta$-TCVAE \citep{mig}. Concerning the former approaches, it has been observed that training InfoGANs is difficult because of the training instability of the GAN framework \citep{betavae}. On the other hand, approaches based on VAEs offer training stability \citep{betavae}, but a recent extensive empirical evaluation has found that their disentangling performance was not reliable as it varied widely with hyperparameter selection, random seeds and among datasets \citep{locatello}.\\

With respect to unsupervised feature extraction, principal component analysis (PCA) is a standard methodology. More in general, it is well established that kernel methods are stable to train and give reliable performance, which contrasts sharply with the issues outlined in the above paragraph. However, in contrast to deep networks, they are shallow methods, so they cannot take advantage of depth to hierarchically decompose a difficult target function into a composition of simpler functions \citep{bfc}. Finally, currently there is no widely accepted technique to promote disentanglement in kernel methods.\\

This paper introduces \arnold, a deep kernel method for the unsupervised learning of disentangled representations. Our approach is based on deep restricted kernel machines (deep RKMs) \citep{drkm}, which provide a framework rooted in well-understood and reliable kernel methods. There has been little investigation on the use of deep RKMs for unsupervised learning. Moreover, while recent evidence suggests that shallow restricted kernel machines can achieve good disentangling performance \citep{pandey2020a}, no previous study has investigated the potential for unsupervised learning of disentangled representations in deep RKMs.\\

We propose employing a deep restricted kernel machine consisting of multiple kernel PCA layers augmented by orthogonality constraints on the latent variables to perform unsupervised extraction of disentangled features. We explain that the orthogonality constraints have two effects, called \textit{intraorthogonality} and \textit{interorthogonality} effects, which encourage the deep RKM to learn a disentangled representation of the data. As a result of the introduction of the orthogonality constraints, we are able to avoid defining a stabilized version of the deep RKM objective, which is instead needed in the original formulation of deep RKMs \citep{drkm}.
Furthermore, we show how to train the proposed model end-to-end, so that the components of lower layers can use the representations learned by higher layers. 
To evaluate the proposed methodology, we conduct experiments in the task of disentangled feature learning. We quantitatively show that RKMs can benefit from depth and that a 2-layer \arnold architecture performs similarly to the state of the art on a number of disentanglement metrics when only a fraction of training points are available, while being more reliable in terms of sensitivity to hyperparameter selection than $\beta$-VAE.\\

Our main contributions are as follows.
\begin{itemize}
\item We propose a novel deep kernel method called \arnold by reformulating the deep restricted kernel machine framework for kernel PCA \citep{drkm} into a constrained optimization problem with orthogonality constraints on the latent variables so that disentanglement is encouraged and optimization can be carried out without first defining a stabilized objective.
\item We propose an end-to-end training algorithm, allowing lower layers to exploit the representation learned by higher layers, by employing a quadratic penalty optimization algorithm with warm start, additionally introducing a deterministic initialization scheme. Furthermore, we present a denoising procedure for deep RKMs.
\item We show how to apply our proposed method to the unsupervised learning of disentangled representations without any prior knowledge on the generative factors, demonstrating empirically that its learned representations perform similarly overall compared to $\beta$-VAE in terms of multiple disentanglement metrics on four benchmark datasets when few training points are available. We also evaluate and discuss the influence of hyperparameters on the performance of the proposed method, demonstrating that our approach is less sensitive to hyperparameter choice than $\beta$-VAE: \arnold's disentangling performance tends to remain steady as its hyperparameter $\gamma$ varies, in contrast to the strong influence of the hyperparameter $\beta$ on $\beta$-VAE's performance. In particular, we show that deterministic initialization of \arnold's training algorithm considerably improves the reproducibility of the results. 
\item Finally, we show the benefit of \arnold over kernel PCA in denoising complex 2D data distributions. In addition, we study the influence of each principal component by visualizing the concept learned by each component in every layer. In this way, we illustrate the role of the different layers: the first layer performs lower-level feature detection and focuses on, for example, edges and corners, while the second layer employs those lower-level features and captures more global features, which in turn allow for more accurate reproduction of the details of the original data distribution.

\end{itemize}

\section{Related work}
\label{sec:rel}

A conventional method to extract features in an unsupervised manner is PCA \citep{pca}. The underlying assumption of PCA is that the input high-dimensional observations lie in a lower-dimensional linear subspace, which is indeed similar to the fundamental assumption of representation learning introduced in the previous section. PCA works through a linear transformation that projects the input observations in a new coordinate system, where the direction of maximum variance is called the first principal component, the orthogonal direction with the second highest variance is called the second principal component, and so on. The dimensionality reduction is achieved by considering only the first $s$ principal components. Kernel PCA is an extension of PCA that introduces nonlinearities by mapping the input observations to a higher dimensional feature space using a kernel function \citep{scholkopf1998}. Linear PCA is then performed in that space. Our proposed method builds on multiple layers of kernel PCA and introduces orthogonality constraints on the latent variables to promote disentanglement. In contrast to our method, neither PCA nor kernel PCA exploits depth or deals with disentanglement. \\

In deep learning, pre-training each layer of a deep network in an unsupervised fashion before fine tuning the entire network is a common technique. In this context, several works have proposed unsupervised methodologies that learn data representations, including \citep{ranzato2007,vincent2008,vincent2010}. For instance, \citep{ranzato2007} proposed an unsupervised encoder-decoder architecture that employed multiple convolutions, sparsity constraints and pooling to build a hierarchical representation of the data and boost invariance; in fact, pooling was shown to represent an information bottleneck that can promote invariance \citep{achille}. Another example is \citep{vincent2010}, where invariant representations were obtained using stacked denoising autoencoders. In that work it was shown that using those representations to train a classifier led to improved  accuracy not only for deep networks, but also for support vector machines with radial basis function kernel, supporting our hypothesis that kernel methods can benefit from deep representations. With the mentioned works, our method shares the idea of exploiting depth, but it applies that idea in the framework of kernel methods instead of neural networks. In particular, differently to \citep{vincent2010}, our method stacks kernel PCA layers instead of autoencoders. Finally, in contrast with the works mentioned in this paragraph, our method takes disentanglement of the learned features into account.\\

One early approach to build disentangled representations was independent component analysis (ICA), which is a method to extract independent components from a signal \citep{comon1994}. It is based on the assumption that the observed signal is a linear mixture of unknown latent signals that are non-Gaussian and mutually independent. However, it was shown empirically that ICA has a poor  disentangling performance compared to the state of the art \citep{betavae}. \\

A more recent approach to the unsupervised learning of disentangled representations is InfoGAN \citep{infogan}. Building upon the GAN framework \citep{gan}, InfoGAN uses a generator $G$ that aims to produce a realistic observation from the concatenation of a noise vector $n$ with a latent representation $z$. Disentanglement in $z$ is encouraged by adding a regularizer to the typical GAN objective. The regularizer aims to maximize a lower bound of the mutual information between $z$ and the generated observation $G(n,z)$. In contrast to InfoGAN, our method is based on kernel methods, which provide more stable training procedures compared to the GAN framework. Finally, InfoGAN's disentangling performance has been empirically shown to be significantly lower than VAE-based methods \citep{betavae}, which represent the state of the art. \\

Most VAE-based methods are based on the $\beta$-VAE approach \citep{betavae}. In this setting, one assumes that the prior over the latent variables $P(z)$ is Gaussian. Then, the posterior $q_\phi(z|x)$, which is approximated using a deep neural network parametrized by $\phi$, is learned by maximizing
\begin{equation} \label{betavae}
L(\theta,\phi;x,z,\beta) = \E_{q_\phi(z|x)} \left[ \log p_\theta(x|z) \right] - \beta D_{\text{KL}}\left( q_\phi(z|x) || p(z) \right), 
\end{equation}
where the likelihood $p_\theta(x|z)$ is approximated by a deep neural network parametrized by $\theta$, $\beta$ is a hyperparameter controlling the degree of disentanglement and $D_{\text{KL}}$ is the Kullbach-Leibler (KL) divergence defined as $D_{\text{KL}}(r(x)|s(x)) = \E_r \left[ \log r/s \right]$. In \citep{betavae} it is claimed that fixing $\beta>1$ promotes disentanglement because it imposes a stricter information bottleneck on the latent factors $z$. \citep{mig} extends this explanation by identifying in Equation \eqref{betavae} a total correlation term that was already known to be related with the disentanglement of a representation \citep{versteeg2015,achille}. Similarly to \citep{mig}, FactorVAE \citep{factorvae} augments the VAE objective (Eq. \eqref{betavae} with $\beta=1$) with an approximation of the total correlation. However, \citep{locatello} performed a large-scale experimental study analyzing six VAE-bases methods, including $\beta$-VAE, $\beta$-TCVAE and FactorVAE, and concluded that they cannot be used to reliably learn disentangled representations in the unsupervised setting, as their disentangling performance was shown to vary widely with hyperparameter selection, random seeds and among datasets.\\ 

Regarding methods based on restricted kernel machines (RKMs) \citep{drkm}, which is the framework our work is built upon, \citep{pandey2020a} has recently proposed a generative model called Gen-RKM, which was experimentally shown to be able to produce disentangled representations. Gen-RKM makes use of a single level of kernel PCA expressed in the RKM framework to derive the latent representations, while the method proposed in this paper introduces a deep architecture made up of multiple kernel PCA objectives.
In fact, while \citep{pandey2020a} considered a single feature map consisting of a deep convolutional neural network, in this paper we consider several feature maps over multiple levels. In other words, following the terminology used in \citep{drkm}, in the former case deep learning is only performed over layers, while in the latter case depth is given by multiple levels, each associated with a different feature map possibly consisting of multiple layers. For the sake of simplicity, in this paper the term ``layer'' is used instead of ``level'' when there is no ambiguity.
Moreover, \arnold introduces new disentanglement constraints on the latent variables, defining in this way a constrained optimization problem that eliminates the need for the stabilization of the loss function used in \citep{drkm} and \citep{pandey2020a} and that can handle explicit and implicit feature maps in the same manner. Before introducing our proposed architecture, it is useful to first briefly illustrate the deep RKM framework.
\section{Background: deep restricted kernel machines}
\label{sec:drkm}

This section reviews the framework of deep restricted kernel machines, as \arnold builds upon it. First, the restricted kernel machine formulation of a single layer of kernel PCA is explained. Then, deep restricted kernel machines are introduced by means of an example. This section ends with the description of an open problem in deep restricted kernel machines: performing end-to-end training and promoting disentanglement at the same time.

\subsection{Restricted kernel machine formulation of kernel PCA}
\label{ssec:kpca}
The restricted kernel machine formulation of kernel PCA gives another expression of the Least-Squares Support Vector Machine (LS-SVM) kernel PCA problem \citep{suykens2003} as an energy with visible and hidden units that is similar to the energy of restricted Boltzmann machines (RBMs) \citep{bengio2009,fischer2014,hinton2006,salakhutdinov2015}. In this new formulation, contrary to RBMs, both the visible units $v_i$ and the hidden units $h_i$ can be continuous. To derive this formulation, consider a training set of $N$ data points of dimension $d$, a feature map $\varphi: \mathbb{R}^d \to \mathbb{R}^{d_F}$ and let $s$ be the number of selected principal components. In the LS-SVM setting, the kernel PCA problem can be written as:
\begin{mini}|l|
	{W,e_i}{J_{\text{kpca}} = \frac{\eta}{2}\mytrace{(W^TW)}-\frac{1}{2\lambda}\sum_{i=1}^Ne_i^Te_i}{}{}
	\label{kpca}
	\addConstraint{e_i}{= W^T\varphi(v_i),}{\quad i=1,\dots,N,}
\end{mini}
where $W \in \mathbb{R}^{d_F \times s}$ is an unknown interconnection matrix, $e_i \in \mathbb{R}^s$ are the error variables and $\eta$ and $\lambda$ are hyperparameters. The restricted kernel machine formulation of kernel PCA \citep{drkm} is given by an upper bound of $J_{\text{kpca}}$ obtained by introducing the latent representations $h_i \in \mathbb{R}^s,\; i=1,\dots,N$ using:
\begin{equation}
\frac{1}{2\lambda}e^Te+\frac{\lambda}{2}h^Th \geq e^Th, \quad \forall e, h \in \mathbb{R}^s.
\end{equation}

This leads to the following objective:
\begin{equation} \label{rkm-kpca}
J_t = -\sum_{i=1}^N \varphi(v_i)^T W h_i+\frac{\lambda}{2}\sum_{i=1}^Nh_i^Th_i+\frac{\eta}{2}\mytrace{\left( W^TW \right)},
\end{equation}
which is the formulation of kernel PCA in the restricted kernel machine framework. The $v_i$ are called visible units because their states are observed, while the hidden units $h_i$ correspond to feature detectors \citep{hinton2012}. In the representation learning literature, $h_i$ is also known as the (latent) representation or code of $v_i$; we say that $h_i$ consists of $s$ latent or code variables or of $s$ hidden features or units. Note that the first term of $J_t$ is similar to the energy of an RBM.\\

As in other energy-based models, the RKM energy for kernel PCA associates a scalar value to each configuration of the variables. As a consequence, given training points $x_i,\;i=1, \dots, N$, training means clamping the visible units $v_i$ to the training points $x_i$ and finding an energy function for which the observed configurations of the visible units are given lower energies than unobserved configurations \citep{lecun2006}. To do so, one characterizes the stationary points of $J_t$, which results in an eigenvalue problem of the kernel matrix $K \in \mathbb{R}^{N \times N}$, with $K_{ij} = \varphi(x_i)^T\varphi(x_j)$ \citep{drkm}.\\

\subsection{Deep restricted kernel machines}
\label{ssec:drkm}
The theory of deep restricted kernel machines was initially proposed in \citep{drkm} with the aim of introducing a new perspective in the connection between deep learning and kernel machines. Such deep RKMs are obtained by coupling several RKMs in sequence, resulting in a deep architecture. An example of a deep RKM is now given.\\

A possible configuration of a deep RKM that extracts features of some data consists of two kernel PCA layers in sequence. Each kernel PCA layer follows the description given in Subsection \ref{ssec:kpca}. The architecture can be summarized in the following manner.
\begin{itemize}
	\item Layer 1 consists of kernel PCA using as input the observation $x_i$ from the given data. The features extracted by this layer are characterized by its hidden features $h_i^{(1)}$.
	\item Layer 2 consists of kernel PCA using as input the hidden features $h_i^{(1)}$ from the preceding layer. The features extracted by this layer are characterized by its hidden features $h_i^{(2)}$.
\end{itemize}
Note that this architecture has a similar structure to stacked autoencoders \citep{bengio2009}: each layer performs unsupervised learning and the hidden features produced by each layer serve as input to the next layer. 
The deep RKM is trained by considering an objective function that joins the objectives of each kernel PCA layer. To formalize the training objective, 
let $s_1$ be the number of selected principal components by the first layer of kernel PCA and $s_2$ be the number of selected principal components by the second layer of kernel PCA.
Then, let $\varphi_1: \mathbb{R}^d \to \mathbb{R}^{d_{\mathcal{F}_1}}$ be the feature map and of layer 1 and 
let $\varphi_2: \mathbb{R}^{s_1} \to \mathbb{R}^{d_{\mathcal{F}_2}}$ be the feature map of layer 2. Also, let $\lambda_1, \lambda_2, \eta_1, \eta_2 > 0$ be hyperparameters. The training objective of the above defined deep RKM is:
\begin{equation}
J_{t,\text{deep}} = J_1 + J_2,
\end{equation}
where $J_1$ and $J_2$ are the objective of a single layer of kernel PCA in the RKM framework as defined in Eq. \eqref{rkm-kpca} using the suitable input. That means:
\begin{equation}
J_1 = -\sum_{i=1}^N \varphi_1(v_i)^T W_1 h_i^{(1)}+\frac{\lambda_1}{2}\sum_{i=1}^N{h_i^{(1)}}^Th_i^{(1)}+\frac{\eta_1}{2}\mytrace{\left( W_1^TW_1 \right)}
\end{equation}
and
\begin{equation}
J_2 = -\sum_{i=1}^N \varphi_2(h_i^{(1)})^T W_2 h_i^{(2)}+\frac{\lambda_2}{2}\sum_{i=1}^N{h_i^{(2)}}^Th_i^{(2)}+\frac{\eta_2}{2}\mytrace{\left( W_2^TW_2 \right)},
\end{equation}
where $W_1 \in \mathbb{R}^{d_{\mathcal{F}_1} \times s_1}$ and  $W_2 \in \mathbb{R}^{d_{\mathcal{F}_2} \times s_2}$ are the interconnection matrices of layer 1 and layer 2, respectively, and $h_i^{(1)} \in \mathbb{R}^{s_1}$ and $h_i^{(2)} \in \mathbb{R}^{s_2}$ are the hidden features of layer 1 and layer 2, respectively.\\

Training the above defined deep RKM means finding the interconnection matrices and the hidden features minimizing the energy $J_{t,\text{deep}}$. Since $J_{t,\text{deep}}$ is unbounded below, to make the energy suitable for minimization, \citep{drkm} proposed to instead minimize a stabilized version of $J_{t,\text{deep}}$. Following \citep{pandey2020a}, this version is defined as:
\begin{equation}
J_{{t,\text{deep}}_{\text{stab}}} = J_{t,\text{deep}} + \frac{c_{\text{stab}}}{2} J_{t,\text{deep}}^2,
\end{equation}
where $c_{\text{stab}} > 0$ is a hyperparameter. It can be shown that $J_{t,\text{deep}}$ and $J_{{t,\text{deep}}_{\text{stab}}}$ share the same stationary points \citep{pandey2020a}.\\

\subsection{Effective algorithms for deep RKMs: an open problem}
Deriving effective algorithms for training deep RKMs is an open problem. In \citep{drkm} a layer-wise training procedure was proposed for the case of linear kernels. However, previous research \citep{bfc} has stressed the importance of end-to-end training in deep architectures to produce representations able to efficiently approximate the target function. The ability of deep learning methods to produce such efficient representations is often linked to hierarchical learning, because deep networks can hierarchically decompose a difficult target function into a composition of simpler functions \citep{bfc}. This ability has been recently explained by a mechanism called ``backward feature correction'' \citep{bfc}, which means that layers of lower abstraction can use the representations learned by layers of higher abstraction to improve the quality of their representation. ``Backward feature correction'' and thus hierarchical learning cannot be achieved using layer-wise training alone \citep{bfc}.\\

On the other hand, it is interesting to note  that training a deep RKM as the one described in Subsection \ref{ssec:drkm} in a layer-wise manner yields mutually orthogonal hidden features, as they are obtained solving an eigenvalue problem of the symmetric kernel matrix $K$. This is an advantage when it comes to the disentanglement of the produced representations, as experimentally shown in \citep{pandey2020a} in the single layer case. However, as it was explained in the previous paragraph, layer-wise training does not take full advantage of the deep architecture, so one would like to perform end-to-end training instead. In the end-to-end training case, one cannot simply solve two eigenvalue problems in sequence, so the mutual orthogonality of the hidden features is lost. Likely, this loss would affect negatively the disentanglement of the learned representations. In this context, the following section proposes \arnold, which is a method that allows to perform end-to-end training and to promote disentanglement at the same time.
\section{The \arnold method}
\label{sec:met}
In the previous section, two key issues in the development of effective training algorithms for deep RKMs were identified: performing end-to-end instead of layer-wise training and promoting disentanglement in the hidden features at the same time. Accordingly, the aim of this section is to propose a method, based on deep RKMs, for the unsupervised learning of latent representations of some given data so that
\begin{itemize}
	\item it promotes disentanglement in the learned hidden features, and
	\item it carries out end-to-end training.
\end{itemize}
To address both aspects, we propose augmenting the original deep RKM formulation of two layers of kernel PCA, as described in Subsection \ref{ssec:drkm}, by orthogonality constraints on the latent variables of both layers.\\

Let $s_1$ be the number of selected principal components of the first layer of kernel PCA and $s_2$ be the number of selected principal components of the second layer of kernel PCA.
Then, let $\varphi_1: \mathbb{R}^d \to \mathbb{R}^{d_{\mathcal{F}_1}}$ be the feature map and of layer 1 and 
let $\varphi_2: \mathbb{R}^{s_1} \to \mathbb{R}^{d_{\mathcal{F}_2}}$ be the feature map of layer 2.
The proposed optimization problem is:
\newcommand{\Jt}{J_{t,\text{\arnold}}}
\newcommand{\Jtb}{\bar{J}{}_{t,\text{\arnold}}}
\newcommand{\Jgbz}{\bar{J}{}_{gen}}
\newcommand{\Jgb}{\bar{J}{}_{gen,\text{\arnold}}}
\begin{mini}|l|
	{W_1,W_2,h_i^{(1)},h_i^{(2)}}{\Jt={\Jt}^{(1)}+{\Jt}^{(2)}}{}{}
	\breakObjective{\quad=-\sum_{i=1}^N \varphi_1(v_i)^T W_1 h_i^{(1)}+\frac{\lambda_1}{2}\sum_{i=1}^N{h_i^{(1)}}^Th_i^{(1)}+\frac{\eta_1}{2}\mytrace{\left( W_1^TW_1 \right)}}
	\breakObjective{\quad\hspace{1.4em}-\sum_{i=1}^N \varphi_2(h_i^{(1)})^T W_2 h_i^{(2)}+\frac{\lambda_2}{2}\sum_{i=1}^N{h_i^{(2)}}^Th_i^{(2)}+\frac{\eta_2}{2}\mytrace{\left( W_2^TW_2 \right)},}
	\addConstraint{\left[
			\begin{array}{c}
				H^{(1)} \\
				H^{(2)}
			\end{array}
			\right]
			\left[
			\begin{array}{cc}
				{H^{(1)}}^T & {H^{(2)}}^T
			\end{array}
			\right]}{= I_{s_1+s_2},}{}
	\label{arnold1}
\end{mini}
where $W_1 \in \mathbb{R}^{d_{\mathcal{F}_1} \times s_1}$ and  $W_2 \in \mathbb{R}^{d_{\mathcal{F}_2} \times s_2}$ are, respectively, the interconnection matrices of layer 1 and layer 2, $h_i^{(1)} \in \mathbb{R}^{s_1}$ and $h_i^{(2)} \in \mathbb{R}^{s_2}$ are, respectively, the latent variables of layer 1 and layer 2, $H^{(1)} = [h^{(1)}_1,\dots,h^{(1)}_N] \in \mathbb{R}^{s_1 \times N}$, $H^{(2)} = [h^{(2)}_1,\dots,h^{(2)}_N] \in \mathbb{R}^{s_2 \times N}$, $I_{s_1+s_2}$ denotes the $(s_1+s_2) \times (s_1+s_2)$ identity matrix and $\lambda_1, \lambda_2, \eta_1, \eta_2 > 0$ are regularization constants.\\

The orthogonality constraints have two effects. The first effect, called the \textit{intraorthogonality} effect, enforces the mutual orthogonality of the hidden features learned by the first layer, as well as the mutual orthogonality of the hidden features learned by the second layer. The second effect, called an \textit{interorthogonality} effect, enforces the orthogonality between the hidden features learned by the first layer and the hidden features learned by the second layer, so that the two layers are encouraged to learn new features of the data instead of repeating the same features in both layers. Both effects aim to push the deep RKM to learn a more disentangled representation of the data. \\

By employing end-to-end instead of layer-wise training in deep restricted kernel machines, lower layers can improve their representation by exploiting the representation learned by higher layers. This architecture may also be beneficial to disentangled feature learning, as it has previously been observed that a two-level hierarchical structure can promote disentanglement \citep{esmaeili2019}. Note that the formulation shown in \eqref{arnold1} can be easily extended to more than two kernel PCA layers. One reason for making use of more layers is that it might improve the invariance of the learned representation, as invariance is promoted by stacking layers \citep{achille}. Another manner of boosting invariance would be to increase the information bottleneck between each layer by selecting fewer principal components \citep{achille}. \\ 

After having trained the model on all training data points $x_i$ and having learned the hidden features $h_i^{(1)}$ and $h_i^{(2)}$ of each $x_i$, one can encode an out-of-sample data point $x^\star$ in the manner proposed in \citep{pandey2020b} extended to the 2-layer case. The latent representation of $x^\star$ is computed by projecting it on the latent space using:
\begin{align}
{h^{(1)}}^\star &= \frac{1}{\lambda_1 \eta_1} \sum_{i=1}^{N} h_i^{(1)} k_0(x_i,x^\star), \label{eq:oosh1}\\
{h^{(2)}}^\star &= \frac{1}{\lambda_2 \eta_2} \sum_{i=1}^{N} h_i^{(2)} k_1(h_i^{(1)},{h^{(1)}}^\star), \label{eq:oosh2}
\end{align}
where $k_0(x,y)=\varphi_1(x)^T\varphi_1(y)$ and $k_1(x,y)=\varphi_2(x)^T\varphi_2(y)$ are the kernel functions for the first and second layer, respectively, obtained by means of the kernel trick. Instead of first defining a feature map $\varphi_i$ and then deriving the kernel, one can simply choose a positive definite kernel $k_{i-1}$ due to Mercer's theorem \citep{mercer1909}, which guarantees the existence of a feature map $\phi$ such that $k_{i-1}(x,y)=\phi(x)^T \phi(y)$.\\

\subsection{A training algorithm for \arnold}
In the energy-based interpretation of deep RKMs, the training phase of \arnold consists of finding the interconnection matrices $W_1$ and $W_2$ and the hidden units $h_i^{(1)}$ and $h_i^{(2)}$ so that the observed configurations $x_i$ are given lower energies than unobserved configurations. Training of \arnold can be therefore performed by solving the equality constrained nonlinear optimization problem \eqref{arnold1}. However, the number of variables of the optimization problem \eqref{arnold1} is large and depends on $d_{\mathcal{F}_1}$ and $d_{\mathcal{F}_2}$, which are the dimensionalities of the feature spaces used by $\varphi_1$ and $\varphi_2$. If, for instance, the Gaussian kernel is used, the dimensionality of the feature space would be infinite and therefore it would not be possible to directly solve the optimization problem \eqref{arnold1}. To address this issue, we show how to rewrite $J_{t,\text{\arnold}}$ to eliminate the interconnection matrices $W_1$ and $W_2$. This can be done considering each layer separately. The following shows how to eliminate the interconnection matrix for the first layer; eliminating $W_2$ follows the same procedure. \\

First, recall that, when training RKMs, each visible unit $v_i$ is fixed to the training point $x_i$. Therefore, in training, the stationary point of $\Jt$ with respect to $W_1$ is given by:
\begin{equation} \label{W1}
\evaluated{\frac{\partial \Jt}{\partial W_1}}_{v_i=x_i} = -\sum_{i=1}^N \varphi(x_i) {h_i^{(1)}}^T + \eta_1 W_1 = 0 \implies  W_1 = \frac{1}{\eta_1} \sum_{i=1}^N \varphi(x_i) {h_i^{(1)}}^T.
\end{equation}

Then, in order to eliminate the interconnection matrix $W_1$, two terms in ${\Jt}^{(1)}$ have to be rewritten: $\sum_{i=1}^N \varphi(v_i)^T W_1 {h_i^{(1)}}^T$ and $\frac{\eta_1}{2}\mytrace{\left(W_1^TW_1\right)}$. This is accomplished using \eqref{W1}. The latter term can be rewritten as follows:

\begin{equation} \label{we_w1}
\begin{split}
\frac{\eta_1}{2}\mytrace{\left(W_1^TW_1\right)} &= \frac{\eta_1}{2}\mytrace{\left(\left(\frac{1}{\eta_1} \sum_{i=1}^N \varphi_1(x_i) {h_i^{(1)}}^T\right)^T \left(\frac{1}{\eta_1} \sum_{i=1}^N \varphi_1(x_i) {h_i^{(1)}}^T\right) \right)} \\
&= \frac{1}{2\eta_1}\mytrace{\left(\left(\sum_{i=1}^N h_i^{(1)} \varphi_1(x_i)^T \right) \left(\sum_{i=1}^N \varphi_1(x_i) {h_i^{(1)}}^T\right) \right)} \\
&= \frac{1}{2\eta_1}\mytrace{\left(\sum_{i=1}^N \sum_{j=1}^N h_i^{(1)} \varphi_1(x_i)^T \varphi_1(x_j) {h_j^{(1)}}^T \right)} \\
&= \frac{1}{2\eta_1}\mytrace{\left(\sum_{i=1}^N \sum_{j=1}^N h_i^{(1)} k_0(x_i,x_j) {h_j^{(1)}}^T \right)} \quad \text{(kernel trick)}  \\
&= \frac{1}{2\eta_1}\mytrace{\left(\sum_{i=1}^N \sum_{j=1}^N k_0(x_i,x_j) h_i^{(1)} {h_j^{(1)}}^T \right)} \\
&= \frac{1}{2\eta_1}\mytrace{\left( H^{(1)}K^{(0)}{H^{(1)}}^T \right)},
\end{split}
\end{equation}
where $K^{(0)} \in \mathbb{R}^{N \times N}$ such that $K^{(0)}_{ij} = k_0(x_i,x_j)$.\\

The former term can be rewritten as follows:
\begin{equation}
\begin{split}
\sum_{i=1}^N \varphi_1(v_i)^T W_1 h_i^{(1)} &= \sum_{i=1}^N \varphi_1(v_i)^T \left( \frac{1}{\eta_1} \sum_{j=1}^N \varphi_1(x_j) {h_j^{(1)}}^T \right) h_i^{(1)} \\
&= \frac{1}{\eta_1} \sum_{i=1}^N \sum_{j=1}^N \varphi_1(v_i)^T \varphi_1(x_j) {h_j^{(1)}}^T h_i^{(1)} \\
&= \frac{1}{\eta_1} \sum_{i=1}^N \sum_{j=1}^N k_0(v_i,x_j) {h_j^{(1)}}^T h_i^{(1)} \quad \text{(kernel trick)} \\
&= \frac{1}{\eta_1} \mytrace{\left( K_{xv}^{(0)} {H^{(1)}}^T H^{(1)} \right)} \\
&= \frac{1}{\eta_1} \mytrace{\left( H^{(1)} K_{xv}^{(0)} {H^{(1)}}^T \right)},
\end{split}
\end{equation}
where $K_{xv}^{(0)} \in \mathbb{R}^{N \times N}$ such that $(K_{xv}^{(0)})_{ij} = k_0(v_i,x_j)$. Note that here $W_1$ was replaced using \eqref{W1} by its expression in terms of the training points $x_i$, but the term $\varphi(v_i)$ was not expressed in terms of the training points. Keeping $\varphi(v_i)$ expressed in $v_i$ is useful when, after training, the visible units are taken as unknowns. For instance, in RBMs this is the case in data generation.\\

Combining the above rewrites, ${\Jt}^{(1)}$ becomes:

\newcommand{\layerbm}[1]{- \frac{1}{\eta_#1} \mytrace{\left( H^{(#1)} K^{(\number\numexpr#1-1\relax)} {H^{(#1)}}^T \right)} + \frac{\lambda_#1}{2} \mytrace{\left( {H^{(#1)}}^T H^{(#1)} \right)} + \frac{1}{2\eta_#1}\mytrace{\left( {H^{(#1)}}K^{(\number\numexpr#1-1\relax)}{H^{(#1)}}^T \right)}}

\newcommand{\layerbmtraining}[1]{- \frac{1}{2\eta_#1} \mytrace{\left( H^{(#1)} K^{(\number\numexpr#1-1\relax)} {H^{(#1)}}^T \right)} + \frac{\lambda_#1}{2} \mytrace{\left( {H^{(#1)}}^T H^{(#1)} \right)}}

\begin{align}
{\Jtb}^{(1)} &= -\sum_{i=1}^N \varphi_1(v_i)^T W_1 {h_i^{(1)}} + \frac{\lambda_1}{2} \sum_{i=1}^N {h_i^{(1)}}^T{h_i^{(1)}} + \frac{\eta_1}{2}\mytrace{\left(W_1^TW_1\right)} \\
&= -\frac{1}{\eta_1} \sum_{i=1}^N \sum_{j=1}^N \varphi_1(v_i)^T \varphi_1(x_j) {h_i^{(1)}}^T {h_i^{(1)}} \\
& \hspace{1.1em} + \frac{\lambda_1}{2} \sum_{i=1}^N {h_i^{(1)}}^T{h_i^{(1)}} + \frac{1}{2\eta_1}\mytrace{\left(\sum_{i=1}^N \sum_{j=1}^N {h_i^{(1)}} \varphi_1(x_i)^T \varphi_1(x_j) {h_i^{(1)}}^T \right)} \\
&= -\frac{1}{\eta_1} \sum_{i=1}^N \sum_{j=1}^N k_0(v_i,x_j) {h_i^{(1)}}^T {h_i^{(1)}} + \frac{\lambda_1}{2} \sum_{i=1}^N {h_i^{(1)}}^T{h_i^{(1)}} \\
& \hspace{1.1em} + \frac{1}{2\eta_1}\mytrace{\left(\sum_{i=1}^N \sum_{j=1}^N k_0(x_i,x_j) {h_i^{(1)}} {h_i^{(1)}}^T \right)} \quad \label{j_t_we1} \\
&= - \frac{1}{\eta_1} \mytrace{\left( H^{(1)} K^{(0)}_{xv} {H^{(1)}}^T \right)} + \frac{\lambda_1}{2} \mytrace{\left( {H^{(1)}}^T H^{(1)} \right)} + \frac{1}{2\eta_1}\mytrace{\left( {H^{(1)}}K^{(0)}{H^{(1)}}^T \right)}.
\end{align}

The optimization problem of \arnold in \eqref{arnold1} can finally be rewritten by combining the expressions for the first and second layer after weight elimination. Given that in training $K_{xv}^{(0)} = K^{(0)}$, as the visible units $v_i$ are clamped to the training points $x_i$, one can write:
\begin{equation} \label{arnold2}
\begin{alignedat}{3}
&\stackunder{minimize}{$h_i^{(1)},h_i^{(2)}$}       &\qquad& \Jtb={\Jtb}^{(1)}+{\Jtb}^{(2)}         & \\
&                  &      & =\layerbm{1}                           &\\
&                  &      & \hspace{1.1em} \layerbm{2}                            &\\
&                  &      & =\layerbmtraining{1}                           &\\
&                  &      & \hspace{1.1em} \layerbmtraining{2}                            &\\
&\text{subject to} &      & \left[
							\begin{array}{c}
							H^{(1)} \\
							H^{(2)}
							\end{array}
							\right]
							\left[
							\begin{array}{cc}
							{H^{(1)}}^T & {H^{(2)}}^T
							\end{array}
							\right] = I_{s_1+s_2}.                     &
\end{alignedat}
\end{equation}

Compared to \eqref{arnold1}, the above optimization problem does not have the interconnection matrices as variables. This means that not only training is more efficient but also that \arnold can deal with explicit and implicit feature maps in the same manner, even if their feature space has infinite dimensionality. In addition, as a consequence of the introduced constraints and of the elimination of the interconnection matrices, it is not necessary to define a stabilized version of the objective to make it suitable for minimization, as it was instead needed in the original formulation of deep RKMs as explained in Subsection \ref{ssec:drkm}. \\

End-to-end training of the proposed deep RKM can be performed by solving the equality constrained nonlinear optimization problem \eqref{arnold2}. The constraint set of \eqref{arnold2} is a Stiefel manifold $\text{St}(s_1+s_2,N)$, so one of the algorithms that have been proposed for optimization on the Stiefel manifold could be employed. For example, one could use the Newton's method on the Stiefel manifold developed in \citep{smith1993,smith1994,edelman1998}. This algorithm generates points along the geodesic, which is expensive to compute because it uses matrix exponentials. Alternatively, one could avoid computing the geodesic by instead exploiting the Cayley transform to determine the search curve, such as in the algorithms proposed in \citep{wen2013,zhu2017}. However, these methods could also be computationally heavy because the complexity of determining the search curve at each iteration is dominated by a matrix inversion.\\

To avoid expensive matrix computations, we propose applying a quadratic penalty optimization algorithm with warm start based on the description given in \citep{nocedal2006}. First, define a function that combines the objective of \eqref{arnold2} with an additional term penalizing solutions violating the orthogonality constraints. One such function is called the quadratic penalty function $Q(h_i^{(1)},h_i^{(2)};\mu)$ and is defined as
\begin{equation} \label{Q}
Q(h_i^{(1)},h_i^{(2)};\mu) = \Jtb + \frac{\mu}{2} \norm{\left[
	\begin{array}{c}
	H^{(1)} \\
	H^{(2)}
	\end{array}
	\right]
	\left[
	\begin{array}{cc}
	{H^{(1)}}^T & {H^{(2)}}^T
	\end{array}
	\right]-I_{s_1+s_2}}^2_F,
\end{equation} 
where $\mu$ is a penalty parameter penalizing violations of the orthogonality constraints. The constrained optimization problem is then replaced by a sequence of unconstrained ones with increasing $\mu$. In the $k$-th unconstrained problem, the $h_i^{(1)}$  and $h_i^{(2)}$ minimizing $Q(h_i^{(1)},h_i^{(2)};\mu_k)$ are sought, with starting point set to the minimizers found in the $(k-1)$-th problem. Initialization can be done with random values drawn from the standard normal distribution or with deterministic layer-wise kernel PCA initialization, such that the initial hidden features of each layer are computed locally using kernel PCA. The unconstrained problems can be then solved by some unconstrained minimization algorithm; in our experiments, we used Adam \citep{adam}. If these subproblems are solved inexactly, in general the quadratic penalty optimization algorithm will not converge to the global solution of \eqref{arnold2}. In practice, we halt the outer loop of Algorithm \ref{alg:train} after a fixed number of iterations that we choose so that it is larger when the number of variables of the optimization problem is higher. This choice translates to running more outer iterations when the number of training point $N$ is higher, as the number of variables depends on $N$.\\

\begin{algorithm}[h]
	\caption{Algorithm for training a 2-layer \arnold using a quadratic penalty optimization algorithm with warm start. ${(h_i^{(1)})}^s_k$ denotes the starting point $h_i^{(1)}$ at iteration $k$. Note that $Q$ also depends on the hyperparameters $\lambda_1, \lambda_2, \eta_1, \eta_2$ and on the kernel functions $k_0$ and $k_1$, which need to be chosen before optimization.} \label{alg:train}
	\begin{algorithmic}[1]
		\Function{Train}{${(h_i^{(1)})}^s_0$, ${(h_i^{(2)})}^s_0$, $\mu_0>0$, $\tau_0>0, p > 1$}
		\For{$k \gets 0, 1, 2, \dots$}
		
		\State $h_i^{(1)} \gets {(h_i^{(1)})}^s_k$
		\State $h_i^{(2)} \gets {(h_i^{(2)})}^s_k$
		
		\Repeat
		\State Update $\{h_i^{(1)}, h_i^{(2)}\} \gets$ \Call{ADAM}{$Q(h_i^{(1)}, h_i^{(2)}; \mu_k)$}
		\Until{$\norm{\nabla Q(h_i^{(1)},h_i^{(2)};\mu_k)} \leq \tau_k$}
		
		\State $\tau_{k+1} \gets \tau_k / 2$
		\State $\mu_{k+1} \gets p \ast \mu_k$
		\State ${(h_i^{(1)})}^s_{k+1} \gets h_i^{(1)}$
		\State $ {(h_i^{(2)})}^s_{k+1} \gets h_i^{(2)}$
		\EndFor
		
		\State \textbf{return} $h_i^{(1)}, h_i^{(2)}$
		\EndFunction
	\end{algorithmic}
\end{algorithm}

Regarding kernels' hyperparameters, they might also be optimized together with the latent variables by adding it as a variable of the optimization problem \eqref{arnold2}. Alternatively, kernels' hyperparameter selection can be carried out by fixing a validation set and selecting the hyperparameter values that perform best on it. When it comes to the scalability of solving \eqref{arnold2}, the size of both $K^{(0)}$ and $K^{(1)}$ grows quadratically in $N$. On the other hand, their size does not depend on the dimensionality $d$ of the input space. Regarding the unknown matrices $H^{(1)}$ and $H^{(2)}$, their size does not depend on $d$, but it depends on $N$, as well as on the number of selected principal components $s_1$ and $s_2$, respectively. In addition, $K^{(1)}$ is computed from $H^{(1)}$, so optimization might suffer from the increased non-linearity.\\

\subsection{Denoising and \arnold}
\label{met:den}
This section presents a reconstruction procedure that can denoise a test point $x^\star$. As in PCA, denoising is carried out by keeping only the first $s$ principal hidden features because one can assume that noise is concentrated in the components of lower variance. Performing reconstruction is straightforward in PCA because it is just a basis transformation, but the deep architecture and non-linear feature maps of \arnold pose a greater challenge. In fact, it is possible that a point in the feature space used by $\varphi_1$ does not have a pre-image in the input space. In addition, multiple non-linear mappings have to be taken into account during reconstruction.\\

We adapt the approach proposed in \citep{mika1999,scholkopf1999} in the context of kernel PCA to a 2-layer \arnold. Extending the following procedure to more than two layers is straightforward. Call $\mathcal{F}_1$ the feature space used by $\varphi_1$ and assume that the mapped data points are centered in $\mathcal{F}_1$. First, ${h^{(2)}}^\star$, the latent representation of $x^\star$ characterized by the second layer, is computed following Eq. \eqref{eq:oosh2}. Similarly, ${h^{(1)}}^\star$ is then computed following Eq. \eqref{eq:oosh1}. Employing this representation, only $s_1$ components are kept, discarding the components that are noisier. The reconstruction of $x^\star$ from its projections $z_k, \, k=1,\dots,s_1$ onto the first $s_1$ principal components in $\mathcal{F}_1$ is
\begin{equation}
\label{eq:invpca}
\P_{s_1}\varphi_1(x^\star) = \sum_{k=1}^{s_1} z_k v^k,
\end{equation}
where $v^k \in \mathcal{F}_1$ is the $k$-th principal component. As explained in \citep{scholkopf1998}, $v^k$ lies in the span of $\varphi_1(x_1) \dots \varphi_1(x_N)$, so \eqref{eq:invpca} can be written as
\begin{equation}
\label{eq:invpca2}
\P_{s_1}\varphi_1(x^\star) = \sum_{k=1}^{s_1} z_k \sum_{i=1}^N \alpha_i^k \varphi_1(x_i).
\end{equation}
The denoised point in the input space is computed by finding a point $\hat{x^\star}$ such that $\varphi_1(\hat{x^\star})$ approximates $\P_{s_1}\varphi_1(x^\star)$. To this aim, we minimize
\begin{equation}
\label{eq:r}
r(\hat{x^\star}) = \norm{\varphi_1(\hat{x^\star}) - \P_{s_1}\varphi_1(x^\star)}^2.
\end{equation}
Using \eqref{eq:invpca2}, the above can be rewritten as
\begin{equation}
\begin{split}
r(\hat{x^\star}) &= \norm{\varphi_1(\hat{x^\star})}^2 - 2 \varphi_1(\hat{x^\star}) \cdot \P_{s_1}\varphi_1(x^\star) + \norm{\P_{s_1}\varphi_1(x^\star)}^2 \\
&=  k_0(\hat{x^\star},\hat{x^\star}) - 2 \sum_{k=1}^{s_1} z_k \sum_{i=1}^N \alpha_i^k k_0(\hat{x^\star}, x_i) + \norm{\P_{s_1}\varphi_1(x^\star)}^2. \label{eq:rho1}
\end{split}
\end{equation}
Given that, by Eq. (17) of \citep{scholkopf1998}, 
\begin{equation} \label{eq:projection}
z_k = \sum_{i=1}^N \alpha_i^k k_0(x_i,x^\star),
\end{equation}
that, in the LS-SVM formulation of kernel PCA, $\alpha_i = 1/\lambda_1 \, e_i$ \citep{suykens2003,suykens2002} and that $e_i = \lambda_1 h^{(1)}_i$ in restricted kernel machines \citep{drkm}, one can rewrite \eqref{eq:rho1} in terms of the hidden units as
\begin{equation}
\label{eq:rho2}
r(\hat{x^\star}) =  k_0(\hat{x^\star},\hat{x^\star}) - 2 \sum_{i=1}^N \beta_i k_0(\hat{x^\star}, x_i) + \norm{\P_{s_1}\varphi_1(x^\star)}^2,
\end{equation}
where
\begin{equation}
\begin{split}
\beta_i &= \sum_{k=1}^{s_1} z_k \alpha_i^k
= \sum_{k=1}^{s_1} \left( \sum_{j=1}^{N} \alpha_j^k k_0(\hat{x^\star}, x_j) \right) \alpha_i^k
= \sum_{j=1}^{N} \sum_{k=1}^{s_1} \alpha_j^k \alpha_i^k k_0(\hat{x^\star}, x_j)
= \sum_{j=1}^{N} \sum_{k=1}^{s_1} (h_j^{(1)})_k (h_i^{(1)})_k k_0(\hat{x^\star}, x_j).
\end{split}
\end{equation}
In general, standard gradient descent methods can be employed to minimize \eqref{eq:rho2}; in this case, note that the last term of \eqref{eq:rho2} is independent of $\hat{x^\star}$. In the case of the RBF kernel, \citep{mika1999} proposed the following iteration scheme for $\hat{x^\star}$:
\begin{equation}
\label{eq:xhatit}
\hat{x^\star}_{t+1} = \frac{\sum_{i=1}^N \beta_i \exp \left( -\norm{\hat{x^\star}_t - x_i}^2 / (2\sigma_1^2) \right) x_i}{\sum_{i=1}^N \beta_i \exp \left( -\norm{\hat{x^\star}_t - x_i}^2 / (2\sigma_1^2) \right)}.
\end{equation}
In denoising, the starting point is set to the noisy observation $x^\star$.

\section{Experimental evaluation}
\label{sec:exp}
The goal of the experimental evaluation is to test the feasibility of the proposed method for denoising and to assess the advantage of the \arnold architecture in the task of unsupervised disentangled feature learning.  \\

\subsection{Denoising}
We applied the denoising procedure explained in Section \ref{met:den} to complex 2D synthetic datasets. Each dataset is generated by selecting $3000$ points as a training set and 750 additional points as a validation set. In this set of experiments, the noise $n$ is white Gaussian with zero mean and standard deviation $\sigma_n$ varying among different values. The number of selected components is $s_1=2$ for the first layer and $s_2=1$ for the second layer; the hidden units are initialized in a layer-wise manner with kernel PCA. All $\eta$ and $\lambda$ are set to $1$. The kernel function employed in both layers is the RBF kernel and its bandwidth is selected employing the validation set. \arnold is first trained on the noisy points to find their latent representations and then each $x_i$ is denoised by computing its pre-image minimizing the expression in Eq. \eqref{eq:rho2}. \\

\begin{figure}[h]
	\centering
	\begin{subfigure}[b]{0.28\textwidth}
		\includegraphics[page=1,width=\textwidth]{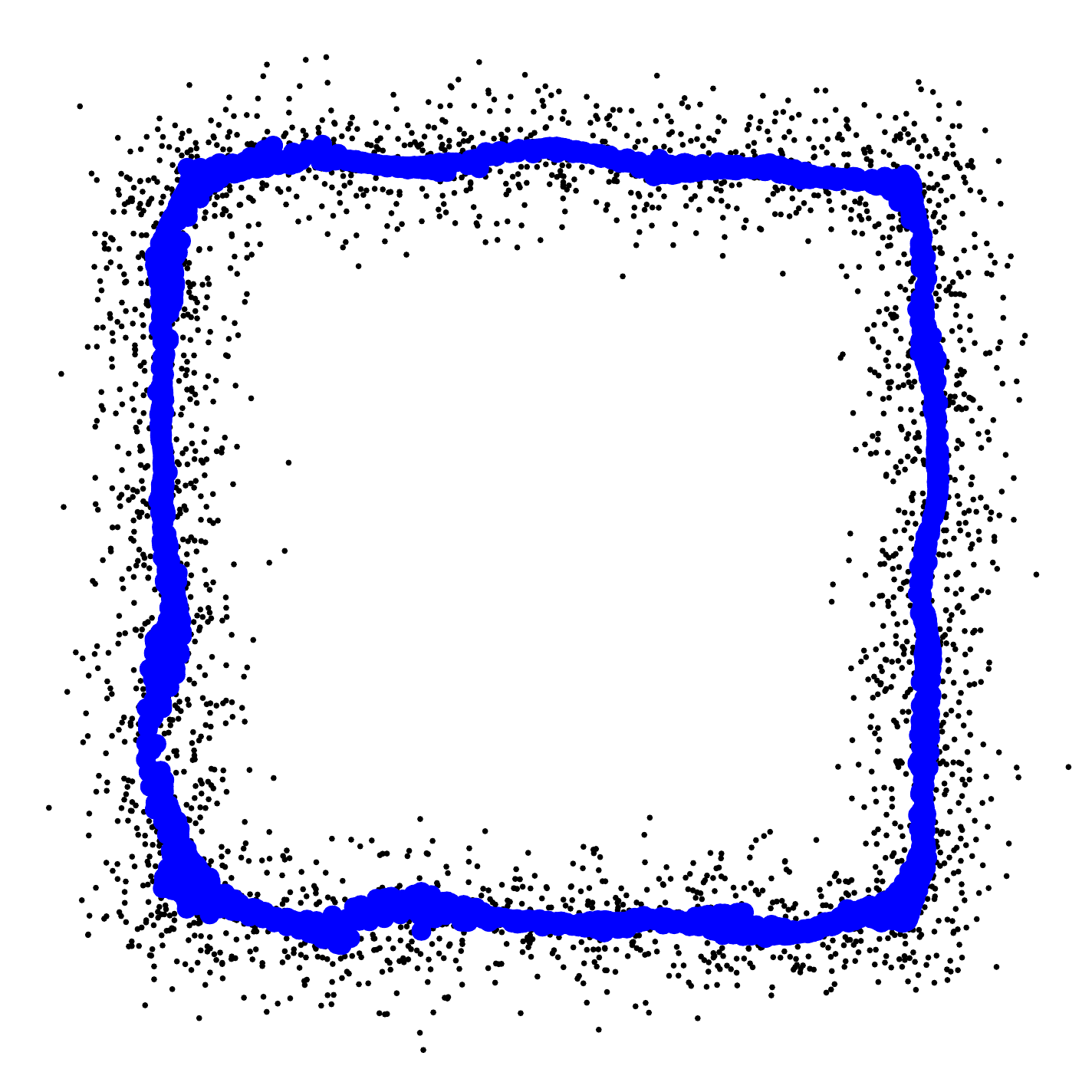}
		\caption{Square.}
		\label{fig:square}
	\end{subfigure}	
	\begin{subfigure}[b]{0.28\textwidth}
		\includegraphics[page=1,width=\textwidth]{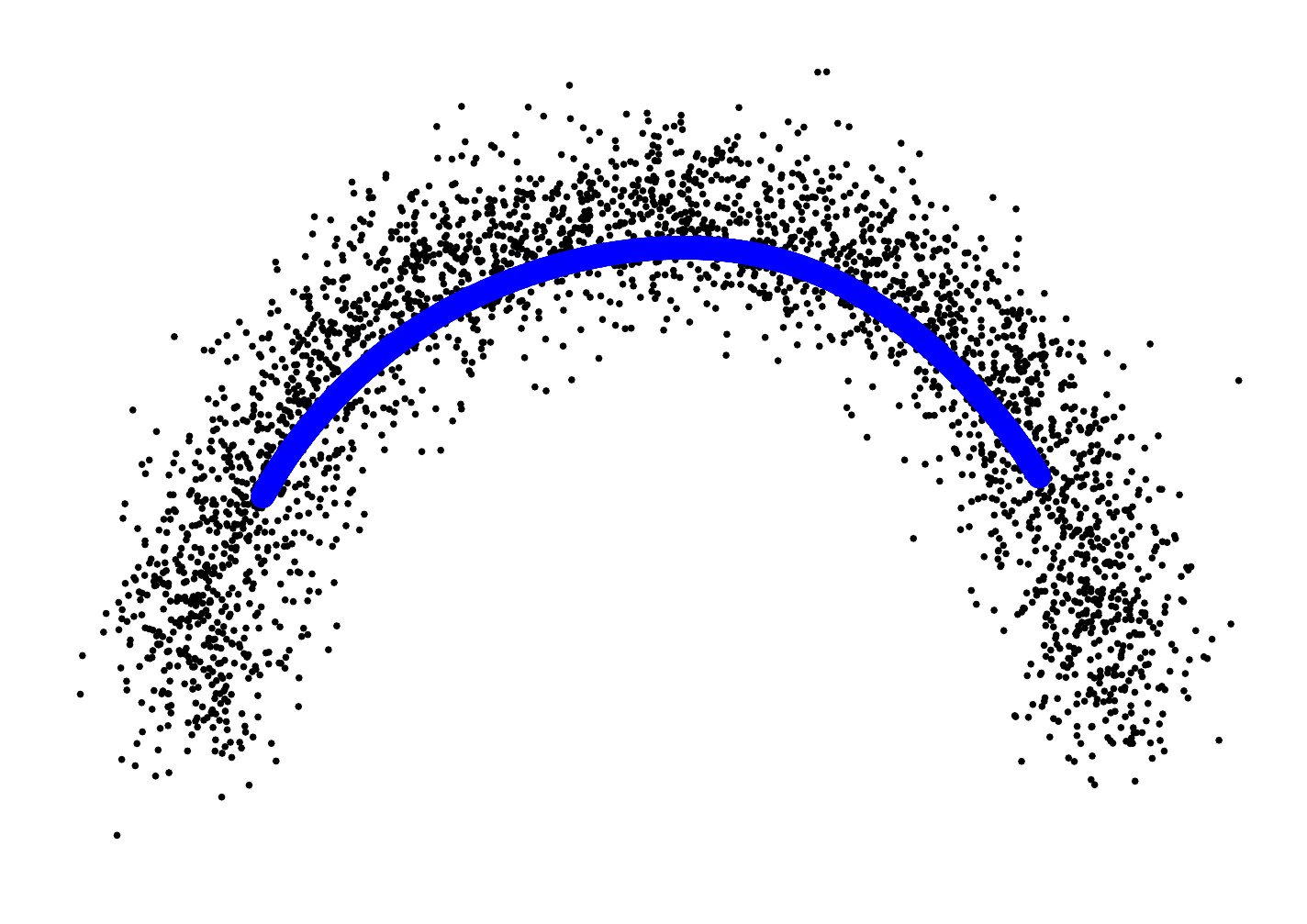}
		\caption{Half circle.}
		\label{fig:halfcircle}
	\end{subfigure}
	
	\caption{Denoising a square and a half circle. The noisy dataset is plotted as smaller black points, its denoised version as larger blue points. The number of selected principal components is $s_1=2$ for the first layer and $s_2=1$ for the second layer. In this experiment, $\sigma_n=0.1$.}
	\label{fig:denoise}
\end{figure}

First, a half circle and a square, depicted in Figure \ref{fig:denoise}, are considered. It can be seen that \arnold successfully captures the structure of the data distributions for both shapes. Note that denoising was effective even though the chosen overall number of principal components was higher than the dimensionality of the datasets. This would not have been the case with linear PCA, which performs perfect reconstruction, hence does not denoise, when using as many components as the dimensionality of the input data.

\begin{figure}[h]
	\centering
	\begin{subfigure}[b]{0.31\textwidth}
		\includegraphics[page=1,width=\textwidth]{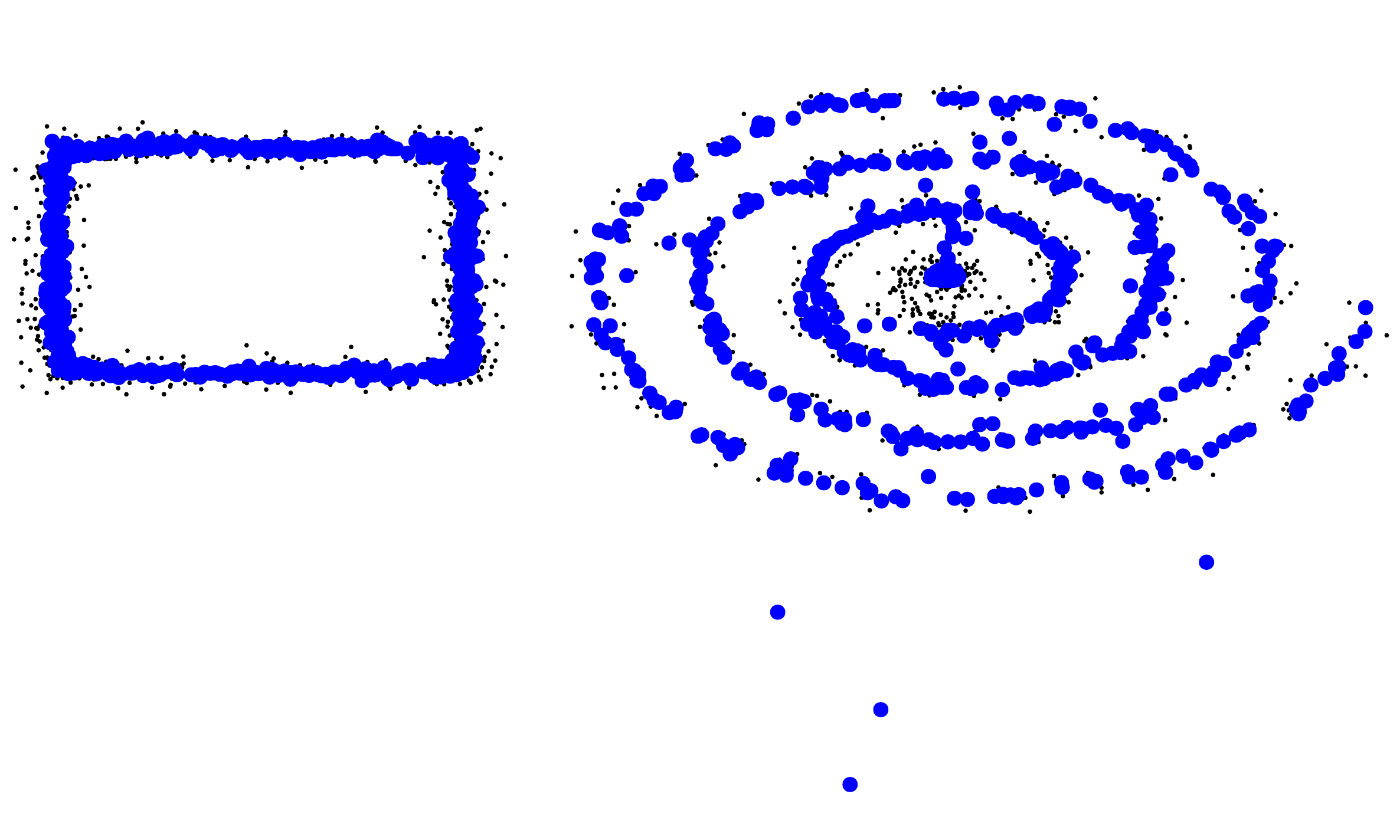}
		\caption{First component, first layer.}
		\label{fig:complexdataset1a}
	\end{subfigure}	
	\hspace{3mm}
	\begin{subfigure}[b]{0.31\textwidth}
		\includegraphics[page=1,width=\textwidth]{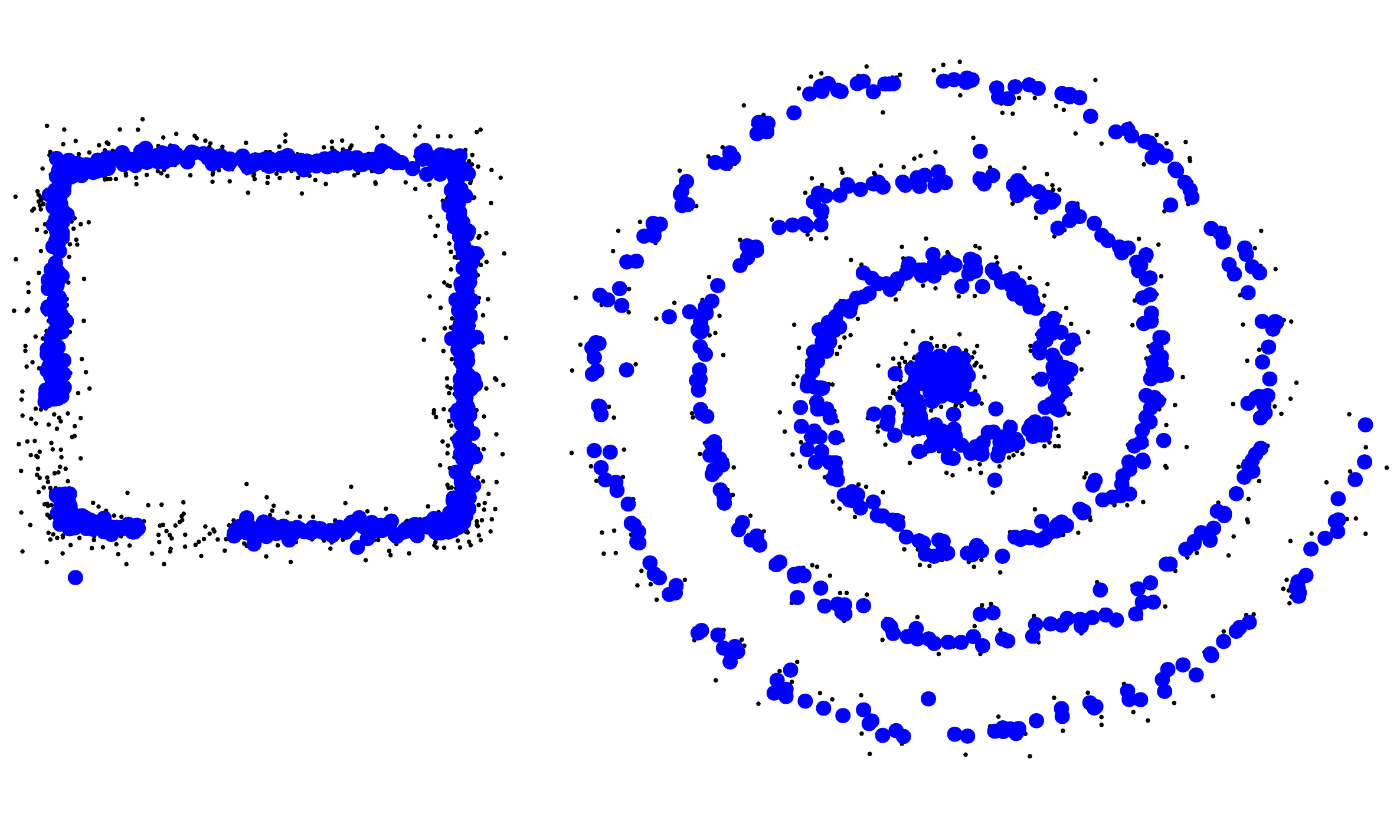}
		\caption{Second component, first layer.}
		\label{fig:complexdataset1b}
	\end{subfigure}
	\hspace{3mm}
	\begin{subfigure}[b]{0.31\textwidth}
		\includegraphics[page=1,width=\textwidth]{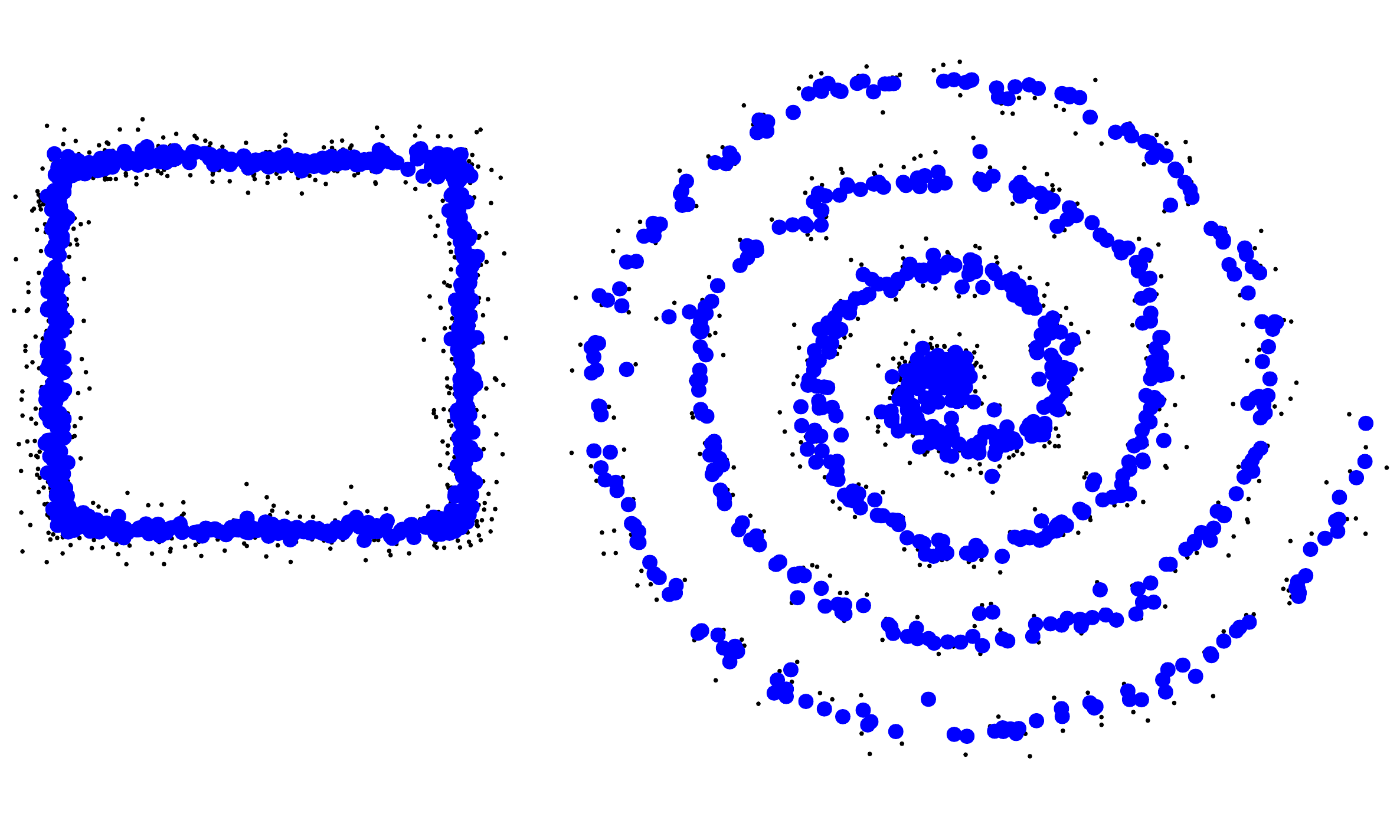}
		\caption{First component, second layer.}
		\label{fig:complexdataset1c}
	\end{subfigure}
	
	\caption{Study of the influence of each component in every layer for a data distribution consisting of a square and a spiral. The number of selected principal components is $s_1=2$ for the first layer and $s_2=1$ for the second layer.}
	\label{fig:complexdataset1}
\end{figure}

Secondly, two additional more complicated datasets were analyzed to study the role of each layer. The result of the influence of each component in every layer can be shown by denoising using only that component. For a dataset with a square and a spiral next to it, three plots, shown in Figure \ref{fig:complexdataset1}, were produced. The first plot is the result of denoising using only the first component of the first layer: it learns the shapes, but has a few outliers and is still noisy around the center of the spiral. The second plot shows the denoised dataset using only the second component of the first layer: it does not show outliers and better reproduces the higher frequency details around the center of the spiral, but loses part of the square. Finally, the third plot is the result of denoising using only the first component of the second layer: it keeps the higher frequency details and reconstructs the square completely. Overall, the principal component of the first layer captures the broad trend but has outliers, while the second component learns the details of the shapes but loses the general trend in some regions. The component of the second layer, on the other hand, both picks up the general trend and reproduces the details of the shapes.\\

\begin{figure}[h]
	\centering
	\begin{subfigure}[b]{0.43\textwidth}
		\includegraphics[page=1,width=\textwidth]{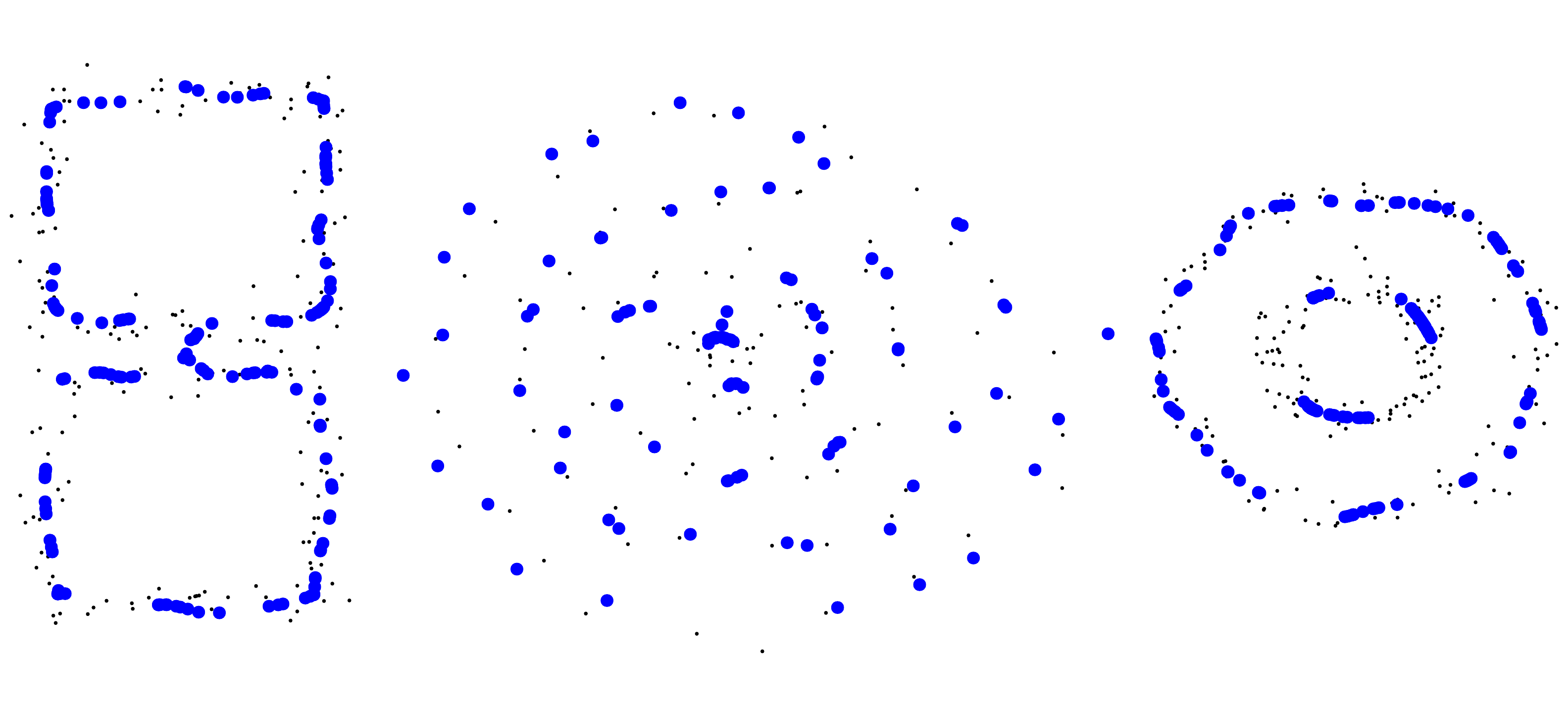}
		\caption{First component, first layer.}
		\label{fig:complexdataset2a}
	\end{subfigure}	
	\hspace{2.5mm}
	\begin{subfigure}[b]{0.43\textwidth}
		\includegraphics[page=1,width=\textwidth]{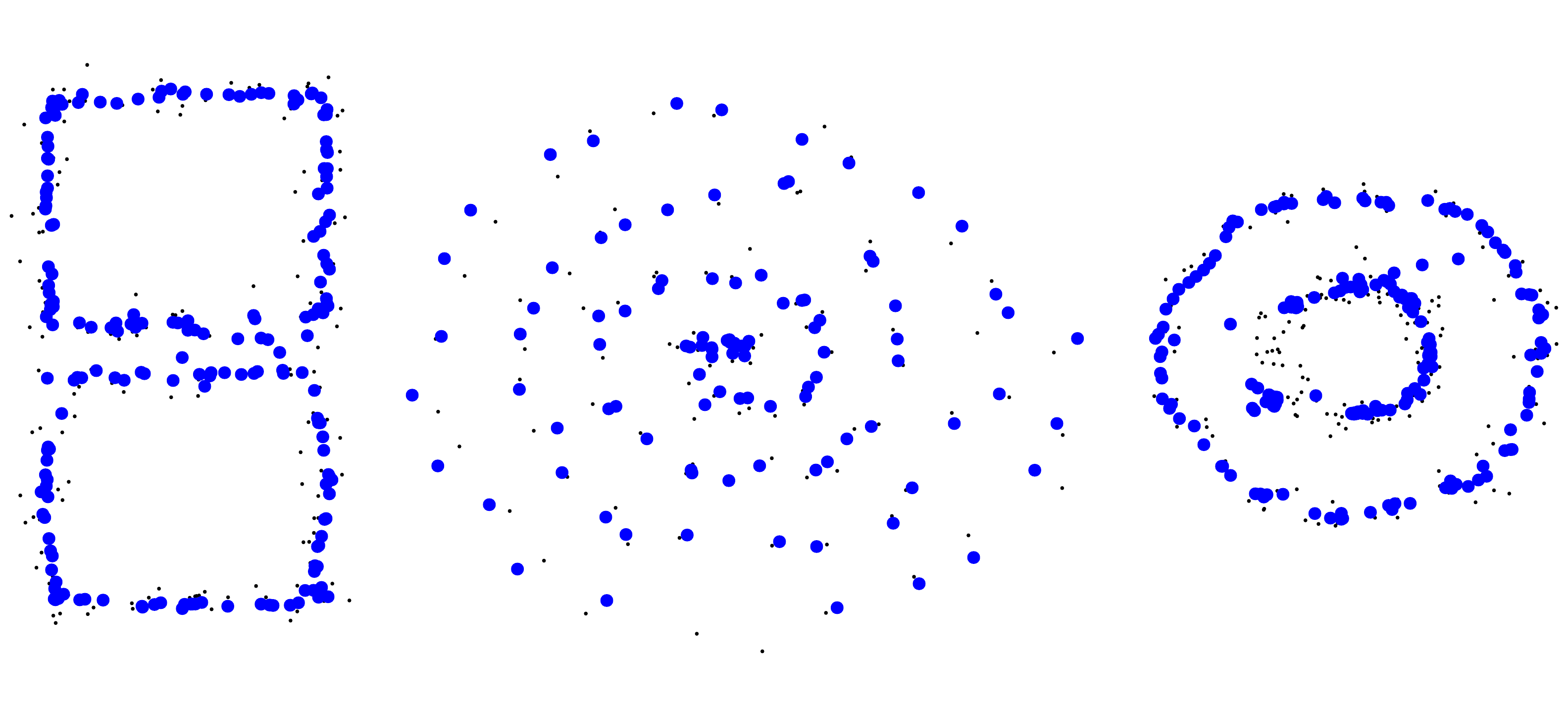}
		\caption{Second component, first layer.}
		\label{fig:complexdataset2b}
	\end{subfigure}
	\hspace{2.5mm}
	\begin{subfigure}[b]{0.43\textwidth}
		\includegraphics[page=1,width=\textwidth]{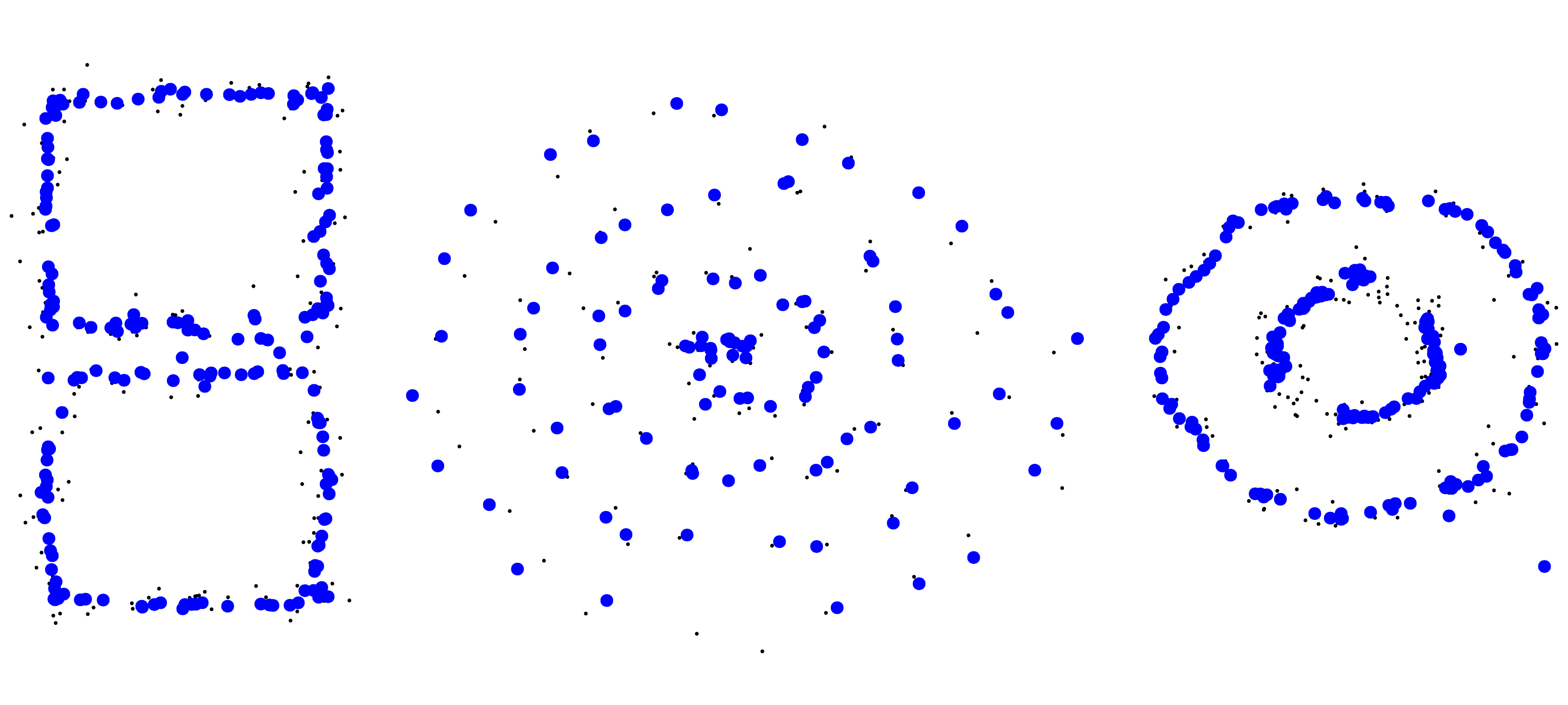}
		\caption{First component, second layer.}
		\label{fig:complexdataset2c}
	\end{subfigure}
	
	\caption{Study of the influence of each component in every layer for a data distribution consisting of two squares, a spiral and a ring. The number of selected principal components is $s_1=2$ for the first layer and $s_2=1$ for the second layer.}
	\label{fig:complexdataset2}
\end{figure}

In addition to the experiment above, a similar analysis for a more complicated data distribution, consisting of two squares, a spiral and a ring, is shown in Figure \ref{fig:complexdataset2}. Using only the first component of the first layer results in some artifacts: two distinct loops of the spiral intersect and two sides of the two squares are joined. On the other hand, the second component reconstructs those shapes correctly, but does not well reproduce the inner circle of the ring. This circle is better reproduced by the first component of the second layer. The findings of the previous two experiments suggest that the lower layer in the deep architecture of \arnold functions as lower-level feature detector focusing on the broad trends of the data distribution, while the higher layer exploits the representation learned by the lower layer and represents higher-level features which lead to better denoising and more accurate reproduction of the original data distribution. These results are consistent with previous findings in deep Boltzmann machines \citep{leroux2008} and in convolutional neural networks \citep{zeiler2014} that lower layers focus on local features, such as edges and corners, and higher layers capture progressively more global and complex patterns with increasing invariance.  \\

Finally, denoising performance was compared to kernel PCA with the same number of overall components. Table \ref{tab:recerr} reports the ratio of the reconstruction error, computed for all denoised points, between \arnold and kernel PCA, where ratios larger than 1 mean that the deep architecture resulted in better denoising than the shallow one. In the considered set of experiments, \arnold outperforms the shallow kernel PCA in denoising. More complicated data distributions, such as the ones in Figure \ref{fig:complexdataset1} and \ref{fig:complexdataset2}, show greater performance gain compared to the shallow architecture. This was expected because kernel PCA is known to be able to effectively denoise simple data distributions \citep{mika1999}, but its denoising performance degrades as the datasets become more complex. The superiority of the deep architecture is strongly marked for small $\sigma_n$, which, together with the observations on the influence of the second layer made in the previous sets of experiments, seems to indicate that the additional layer has a crucial role in the reconstruction of the finer details of the data distribution, as this effect might become less noticeable with higher noise levels. On the whole, our proposed architecture's representational efficiency benefited from depth, since it attained improved denoising of the data distributions in the same number of principal components as the shallow architecture. \\

\begin{table}[h]
	\centering
	\begin{tabular}{lllll}
		\hline
		$\sigma_n$ & Square & Half circle & Dataset of Figure \ref{fig:complexdataset1} & Dataset of Figure \ref{fig:complexdataset2} \\ \hline
		0.05     & 1.22      & 1.36           & 3.18                   & 2.51                   \\
		0.1      & 1.09      & 1.17           & 1.70                   & 1.62                   \\
		0.2      & 1.06      & 1.08           & 1.24                   & 1.21                   \\ \hline
	\end{tabular}
	\caption{Reconstruction error ratios between \arnold and kernel PCA in denoising complex 2D synthetic data distributions for different noise levels. Ratios larger than 1 mean that the deep architecture resulted in better denoising than the shallow one and the larger than 1 the better. In the deep architecture, the number of selected principal components is $s_1=2$ for the first layer and $s_2=1$ for the second layer, while the number of principal components used by kernel PCA is 3.}
	\label{tab:recerr}
\end{table}

\subsection{Disentanglement}
This section aims to quantitatively assess that \arnold is able to learn a disentangled representation of the factors of variation of the data. In addition, the role of the hyperparameters, of the number of selected principal components and of the number of layers is studied. We applied our method on the Cars3D dataset \citep{cars3d}, on the dSprites dataset \citep{betavae} and on the SmallNORB dataset \citep{lecun2004}, as well as on a noisy version of dSprites introduced in \citep{locatello} obtained by replacing the background pixels with Gaussian noise with zero mean and unit variance. In all datasets, each data point is generated according to a deterministic function of its ground-truth latent representation. All data points are $64 \times 64$ images: Cars3D and noisy dSprites contain RGB images, while dSprites and SmallNORB contain grayscale images. For details of the datasets, see \ref{app:datasets}. \\

In the experiments, the dimension of the learned latent representation is fixed to 10: this choice was also made in the large experimental evaluation of \citep{locatello} and, furthermore, this number is greater than but close to the ground-truth number of factor of variations. Computing the hidden features of some data point in \arnold translates to selecting 10 principal components. We chose to do so in the deep architecture with $n_\text{layers} \geq 1$ of \arnold by either fixing $s_i$ to 10 for some $i$ such that $1 \leq i \leq n_\text{layers}$ or by having $\sum_{i=1}^{n_\text{layers}}s_i=10$ and concatenating all $h^{(i)}$. In all experiments, the learned models are evaluated on a subset of $N_\text{eval}=4000$ data points chosen randomly from the relevant dataset. \\

Given that there is no single widely accepted measure to quantify disentanglement, we use three metrics that have been proposed in the literature, namely the IRS score \citep{irs}, the mutual information gap (MIG) \cite {mig} and the SAP score \citep{sap}. The IRS metric measures robust disentanglement, which means that, if a latent variable is associated with some generative factor $G$, the inferred value of that latent variable shows little change when $G$ remains the same, regardless of changes in the other generative factors. The MIG metric is computed as the average over all generative factors of the difference between the two latent variables with highest mutual information with each generative factor. The SAP score is formulated by first building a score matrix $S$ such that $S_{ij}$ is the classification score of predicting the $j$-th generative factor using only the $i$-th latent variable; the final score is the mean of the differences between the top two entries for each column, which corresponds to averaging over the generative factors. For all considered metrics, a higher score indicates better disentangling performance. \\

\begin{figure}[t]
	\centering
	\includegraphics[page=1,width=1.0\textwidth]{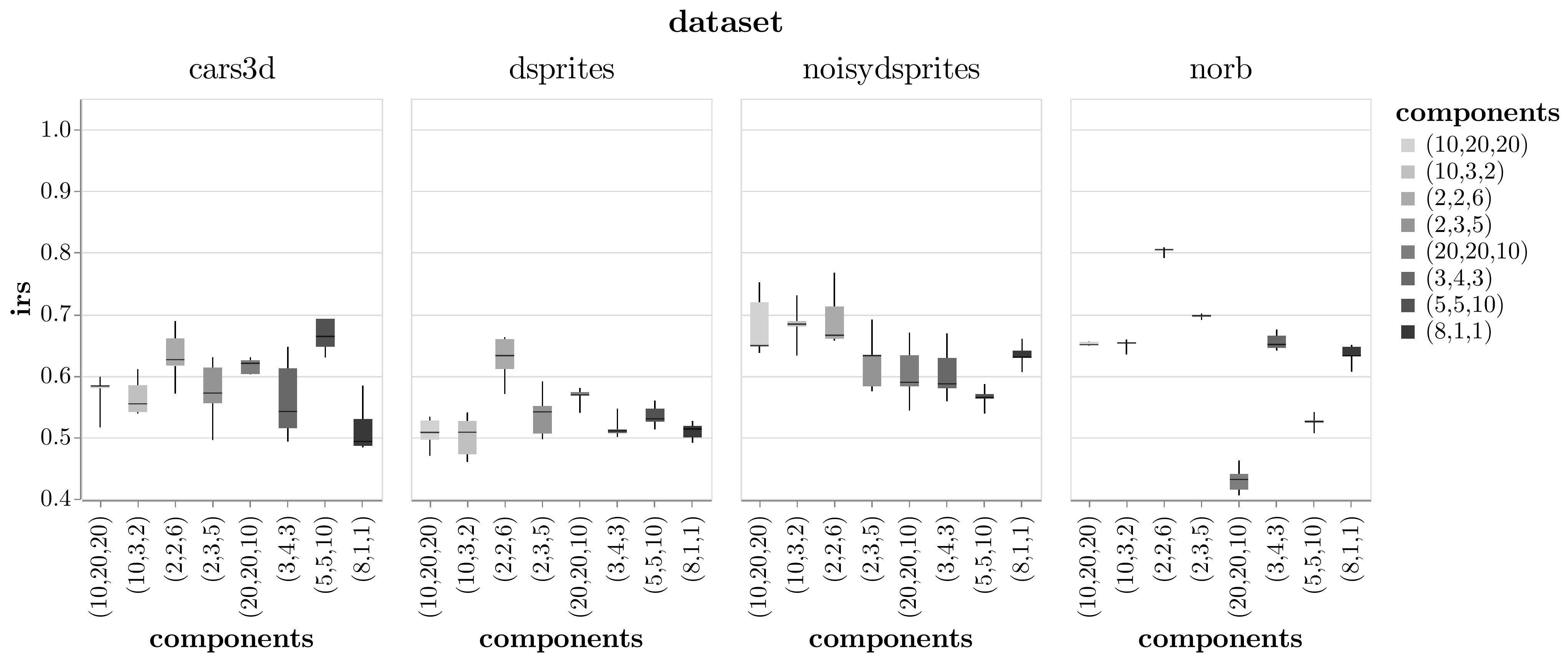}
	\caption{Boxplot of the IRS score of a 3-layer \arnold architecture according to the number of selected principal components for each dataset. The tuples in the labels are of the form $(s_1,s_2,s_3)$. Lower and upper box boundaries first and third quartile, respectively, line inside box median, lower and upper error lines minimum and maximum value, respectively. The results are shown over five random seeds. A higher score indicates better disentangling performance on all metrics.}
	\label{fig:plot8_irs}
\end{figure}

In the first set of experiments, the role of the number of selected principal components is investigated. The studied architecture has $n_\text{layers}=3$. All $\eta$ and $\lambda$ are set to $1$. The chosen kernel function is the RBF kernel and its bandwidth is added as a variable to the optimization problem; this is the case for all experiments in this section. For each dataset, the training set is a random sample of $N=100$ data points. Eight different choices for the number of selected components $s_1$, $s_2$ and $s_3$ are considered. Some representative results are presented in Figure \ref{fig:plot8_irs}. For additional plots, see \ref{app:plots}.\\

Figure \ref{fig:plot8_irs} shows the IRS score attained by \arnold models with varying number of selected principal components in its three layers. The variance is due to five different random seeds. It is clear that the number of selected components has an important role in the disentangling performance of \arnold. Its influence is particularly evident in the SmallNORB dataset, where a bad choice of $s_1,s_2$ and $s_3$ led to considerably worse scores than the other choices. None of the evaluated choices of $s_1,s_2$ and $s_3$ consistently resulted in poor performance on all datasets: for instance, $(20,20,10)$ was not the best performer on the SmallNORB dataset, but it was the choice with the best median score on the dSprites dataset. In a similar manner, no choice of the number of selected components was found to always give higher disentanglement score than any other choice for the datasets considered in the experiments. However, some choices led to better scores more consistently than others, while also showing smaller variance. For instance, models with $s_1=2$, $s_2=2$ and $s_3=6$ never resulted in significantly poorer performance than the other models and always had modest variance with respect to random seeds in the experiments. In general, the variance due to randomness varied depending on the dataset and on the disentanglement metric considered. For instance, the variance for the IRS metric on SmallNORB was small, but it was large for the SAP score metric on the same dataset. Interestingly, one can see that, in general, randomness affects certain combinations of dataset and disentanglement metric more than others. For instance, in Figure \ref{fig:plot8_irs} most models on SmallNORB have small variance, whereas most models have higher variance on the other datasets. \\

\begin{figure}[t]
	\centering
	\includegraphics[page=1,width=0.8\textwidth]{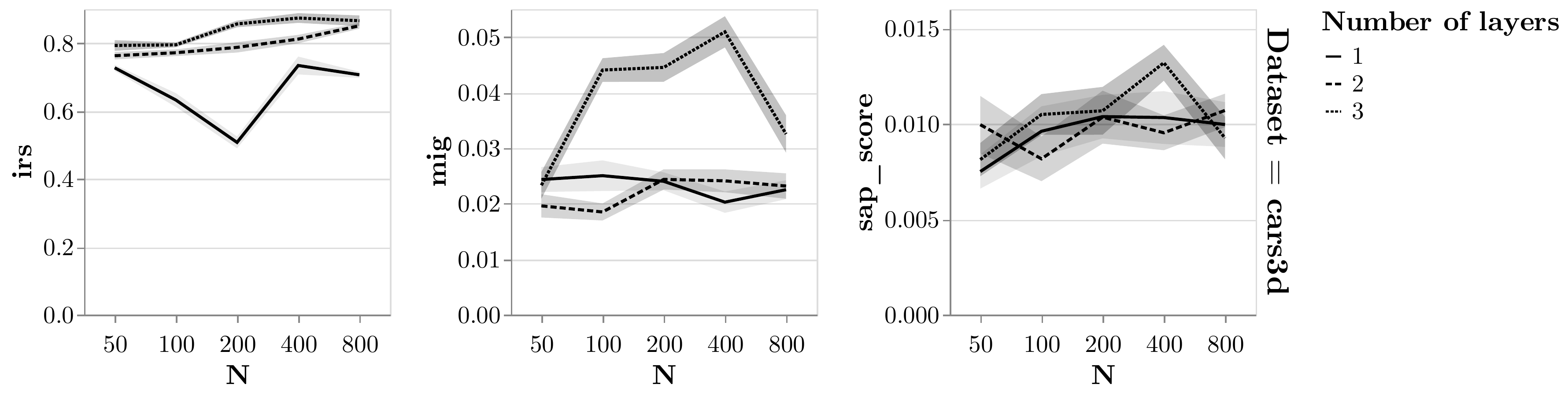}
	\caption{Line chart of the mean disentanglement score of different \arnold architectures according to the number $N$ of training data points and the number $n_\text{layers}$ of layers for the Cars3D dataset. For the 1-layer architecture, $s_1=10$, for the 2-layer one, $s_1=10$ and $s_2=5$ and for the 3-layer one, $s_1=2$, $s_2=2$ and $s_3=6$. The size of each error band is set to the value of standard error, extending from the mean. The variance is due to five different random seeds and different $\gamma$. The higher the curve, the better.}
	\label{fig:plot9_cars3d}
\end{figure}

In the second set of experiments, the role of the number $n_\text{layers}$ of layers on \arnold's disentangling performance is studied as the number $N$ of training points grows from 50 to 800. In the experiments, $n_\text{layers}$ is taken from $\{1,2,3\}$. The studied architectures are as follows: the 1-layer architecture has $s_1=10$, the 2-layer one has $s_1=10$ and $s_2=5$ and for the 3-layer architecture, $s_1=2$, $s_2=2$ and $s_3=6$. For the 2-layer and 3-layer architectures, we chose those configurations because they were among the best ones that were empirically evaluated. On top of varying $N$, the hyperparameters $\eta$ and $\lambda$ are varied as well. Given that they have the role of weights in the objective function, we can consider a single hyperparameter $\gamma = \frac{\eta}{\lambda}$. In the experiments, $\gamma$ is taken from $\{0.01,0.1,1,5,25\}$. All experiments are repeated over five random seeds. The results for the Cars3D dataset are shown in Figure \ref{fig:plot9_cars3d}.\\

Figure \ref{fig:plot9_cars3d} plots the disentanglement score attained by \arnold against the number $N$ of training data points according to $n_\text{layers}$. The results indicate that the extent of the influence of $n_\text{layers}$ on the disentangling performance of \arnold greatly depends on the disentanglement metric. In particular, on Cars3D all considered $n_\text{layers}$ had similar SAP score, while varying the number of layers generally greatly affected the MIG and IRS scores. For example, models with 3 layers resulted in approximately double the MIG score on Cars3D compared to the 2-layer and 1-layer models when $N=400$. From the experiments it can be noted that the most consistent choice in terms of disentangling performance is $n_\text{layers}=2$, as it is in most cases the best or close to the best choice for any considered combination of metric, dataset and $N$. This observation accords with our hypothesis that introducing an additional layer to RKMs can be beneficial in terms of the disentanglement of the learned representation. For the datasets considered in these experiments, adding a third layer did not consistently increase the disentanglement scores, but this may not be the case on more difficult datasets with, for instance, multiple more realistic objects and a complex background. \\

\begin{figure}[h!]
	\centering
	\begin{subfigure}[b]{0.8\textwidth}
		\includegraphics[page=1,width=\textwidth]{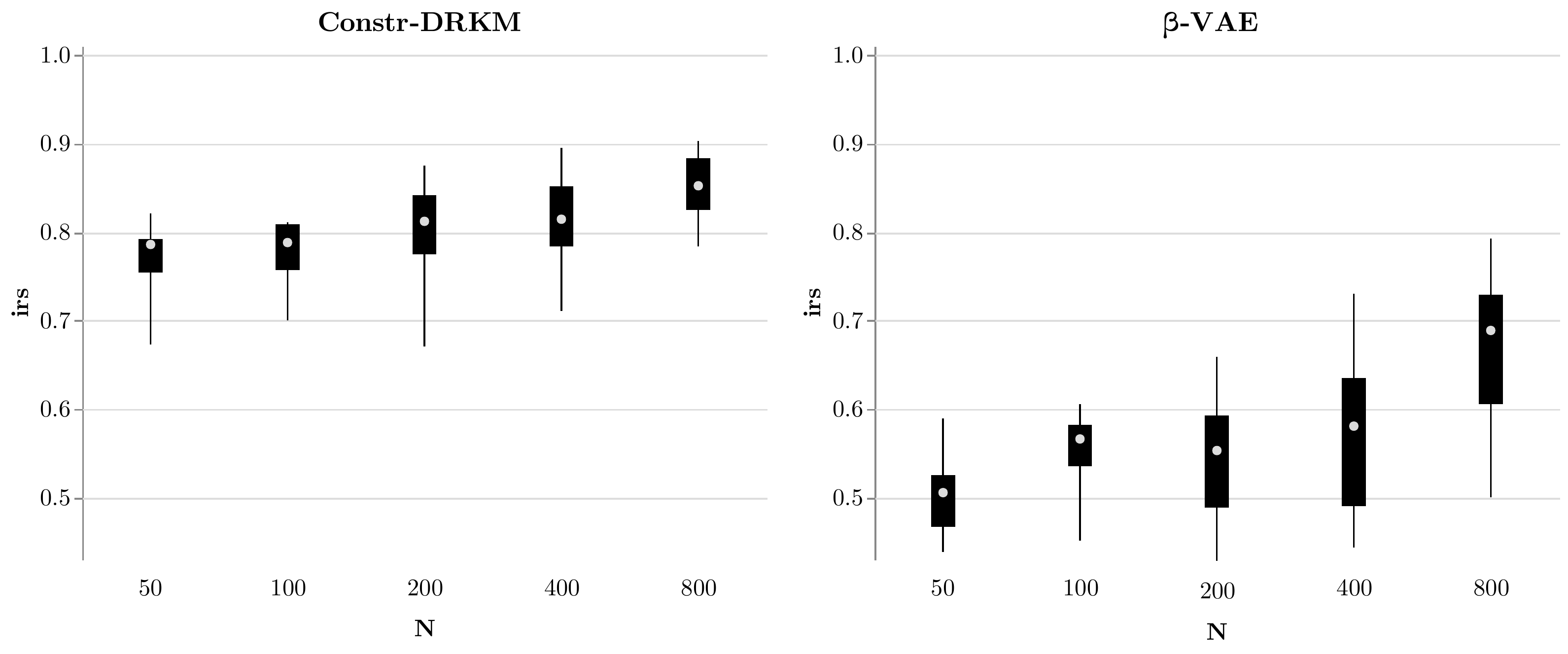}
		\caption{dataset = Cars3D, metric = IRS}
		\label{fig:plot1_cars3d_irs}
	\end{subfigure}

	\begin{subfigure}[b]{0.8\textwidth}
		\includegraphics[page=1,width=\textwidth]{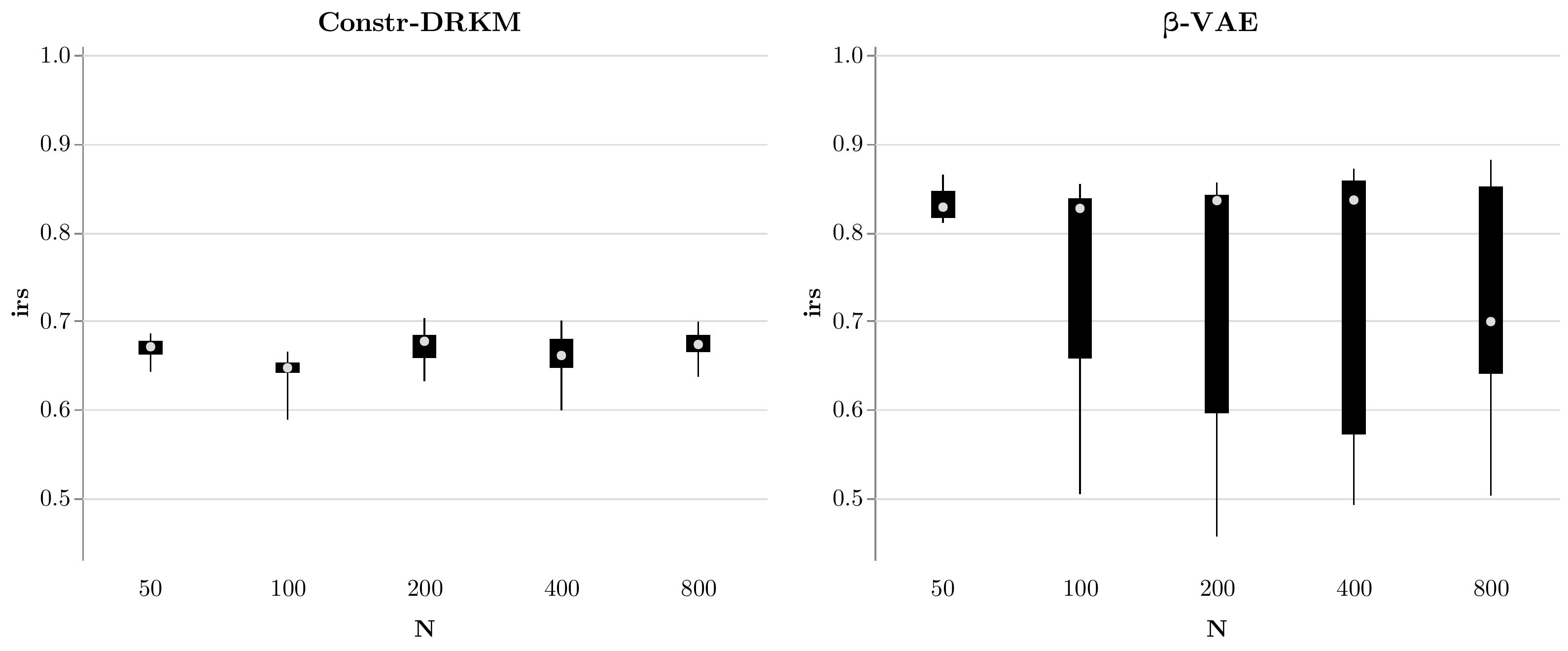}
		\caption{dataset = SmallNORB, metric = IRS}
		\label{fig:plot1_norb_irs}
	\end{subfigure}

	\begin{subfigure}[b]{0.8\textwidth}
		\includegraphics[page=1,width=\textwidth]{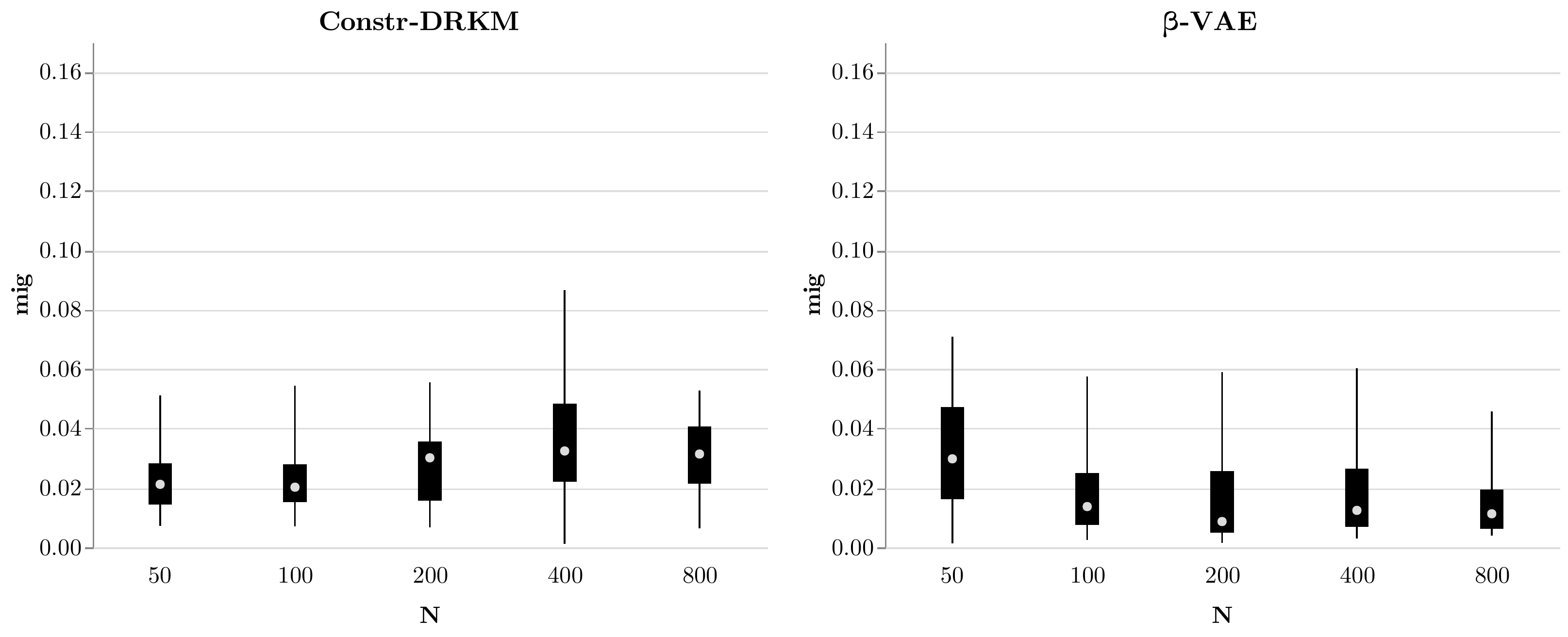}
		
		\caption{dataset = SmallNORB, metric = MIG}
		\label{fig:plot1_norb_mig}
	\end{subfigure}
	
	\caption{Boxplots of the disentanglement score of a 2-layer \arnold, with $s_1=10$ and $s_2=5$, and of a $\beta$-VAE model according to the number of training points on the cars3D and SmallNORB datasets. Disentanglement score are shown across all choices of the hyperparameter $\gamma$ for \arnold and of $\beta$ for $\beta$-VAE and across five random seeds. The boxes show the first and third quartile in the lower and upper box boundaries, respectively, the circle inside is the median and the lower and upper error lines are the minimum and maximum value, respectively. Higher is better on all metrics.}
	\label{fig:plot1}
\end{figure}

In the third set of experiments, the performance of a 2-layer \arnold is studied as the number $N$ of training points grows from 50 to 800 and is compared against $\beta$-VAE \citep{betavae}. The studied architecture has $s_1=10$ and $s_2=5$. The hyperparameter $\gamma$ is varied in the same range as in the previous set of experiments. We repeat the same experiments using a $\beta$-VAE model in place of \arnold where, instead of the hyperparameter $\gamma$, we vary $\beta$ in the set $\{2,3,4,5,6\}$. Some key results are presented in Figure \ref{fig:plot1}. \\

Figure \ref{fig:plot1} plots the disentanglement score attained by both \arnold and $\beta$-VAE against the number of training data points, for a selection of datasets and metrics. The variance is due to different hyperparameters ($\gamma$ for \arnold and $\beta$ for $\beta$-VAE) and five random seeds. Overall, \arnold showed good disentangling performance compared to $\beta$-VAE across datasets and metrics. Remarkably, on Cars3D \arnold significantly outperformed $\beta$-VAE in the IRS score. Turning now to the SmallNORB dataset, \arnold and $\beta$-VAE produced similar MIG scores. On the other hand, the latter method resulted in considerably better median IRS scores than the former method. If we now turn to the analysis of variance, in accordance with \citep{locatello}, $\beta$-VAE's performance varied greatly with random seed and hyperparameter. For example, on SmallNORB in Figure \ref{fig:plot1_norb_irs} the attained score varies from about 0.5 up to almost 0.9 with considerable interquartile range for all $N$ but the smallest. Comparing $\beta$-VAE's variance to \arnold's, it can be seen that our method shows a significantly more limited variance for all $N$. Even on other datasets, \arnold shows small variance on the IRS score that is similar or lower than the one shown by $\beta$-VAE. This is not always the case for other metrics, as exemplified in Figure \ref{fig:plot1_norb_mig}. Overall, the observations made in this set of experiments suggest that \arnold is able to consistently learn a disentangled representation of the input data, competitively with the state of the art, while being less affected by randomness and hyperparameter selection than $\beta$-VAE for the IRS metric. \\

\begin{figure}[h!]
	\centering
	\begin{subfigure}[b]{0.8\textwidth}
		\includegraphics[page=1,width=\textwidth]{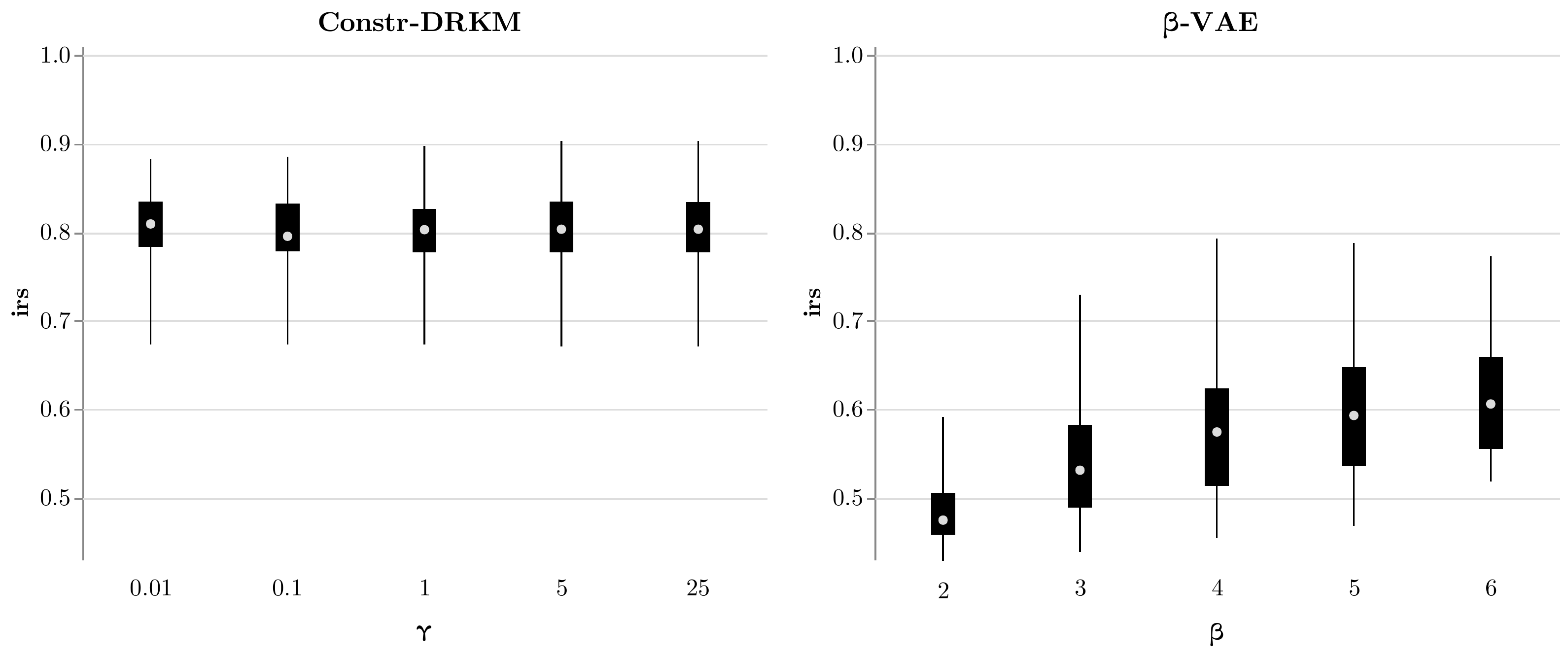}
		\caption{dataset = Cars3D, metric = IRS}
		\label{fig:plot2_cars3d_irs}
	\end{subfigure}

	\begin{subfigure}[b]{0.8\textwidth}
		\includegraphics[page=1,width=\textwidth]{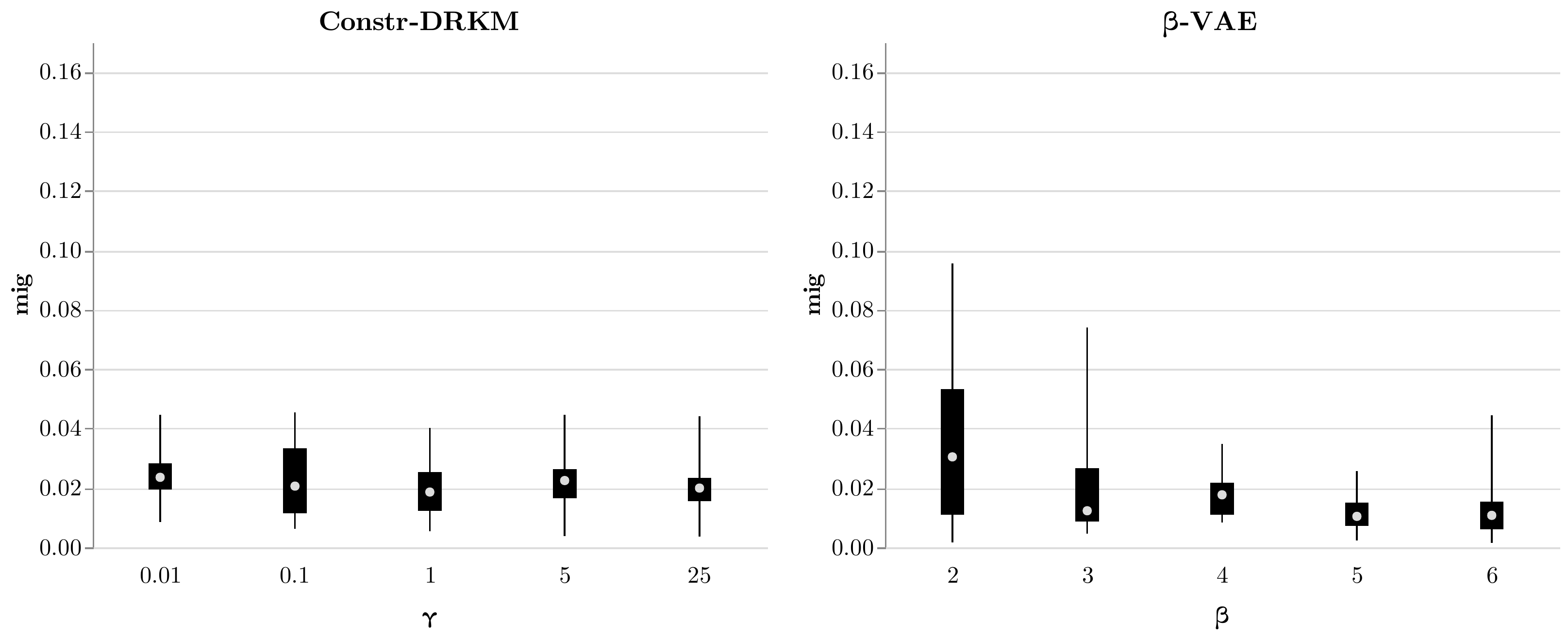}
		\caption{dataset = Cars3D, metric = MIG}
		\label{fig:plot2_cars3d_mig}
	\end{subfigure}

	\begin{subfigure}[b]{0.8\textwidth}
		\includegraphics[page=1,width=\textwidth]{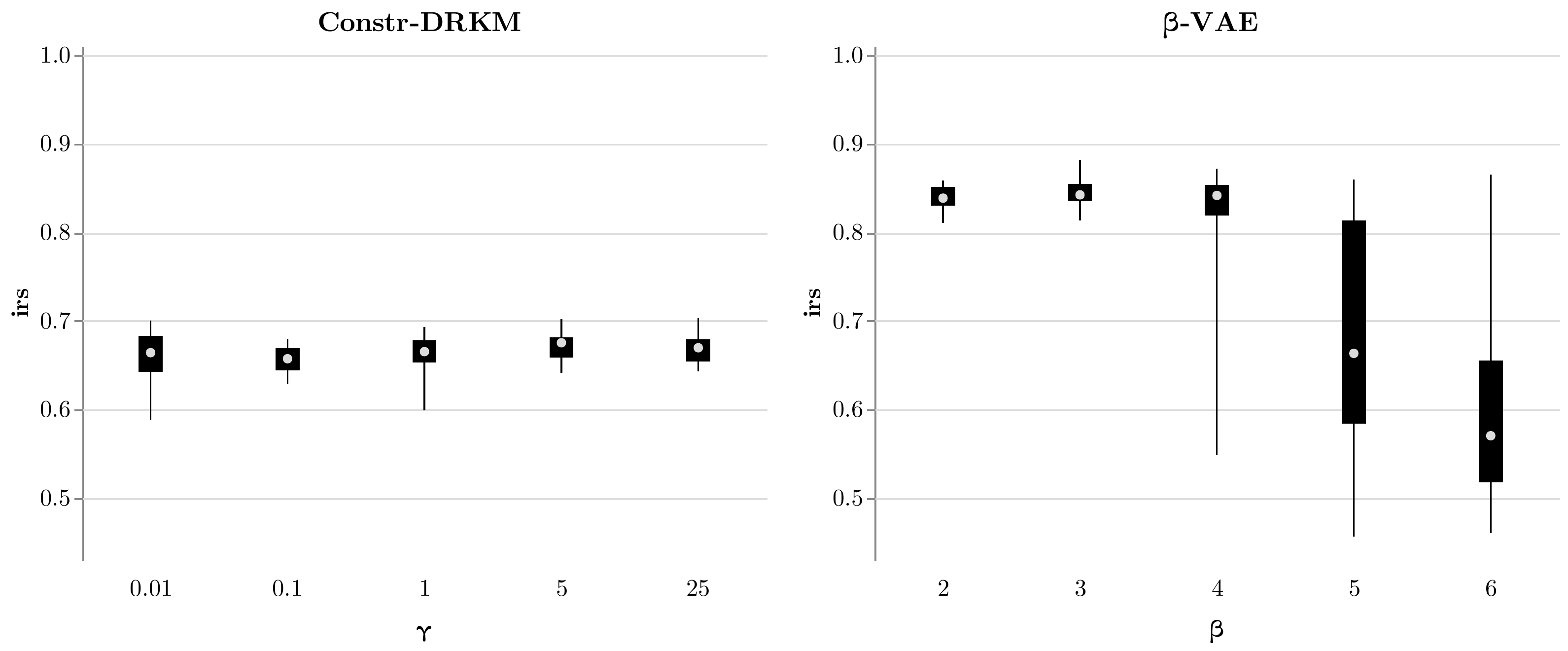}
		\caption{dataset = SmallNORB, metric = IRS}
		\label{fig:plot2_norb_irs}
	\end{subfigure}
	
	\caption{Boxplots of the disentanglement score of a 2-layer \arnold, with $s_1=10$ and $s_2=5$, and of a $\beta$-VAE model according to the hyperparameter $\gamma$ for \arnold and of $\beta$ for $\beta$-VAE on the cars3D and SmallNORB datasets. Disentanglement score are shown across all number $N$ of training points and across five random seeds. The boxes have the same structure as in Figure \ref{fig:plot1}. Higher is better on all metrics.}
	\label{fig:plot2}
\end{figure}

We now focus on the influence of the hyperparameter $\gamma$ for \arnold and $\beta$ for $\beta$-VAE. Figure \ref{fig:plot2} plots the disentanglement score attained by both \arnold and $\beta$-VAE as the hyperparameter $\gamma$ for \arnold and $\beta$ for $\beta$-VAE varies, for a selection of datasets and metrics. The variance is due to different $N$, which is in the same range as in Figure \ref{fig:plot1}, and five random seeds. From Figure \ref{fig:plot2_norb_irs} it can be observed that $\gamma$ does not have appreciable influence on the IRS score on SmallNORB, whereas $\beta$ plays an important role in $\beta$-VAE's median performance, which dropped when setting $\beta$ to 5 and 6, producing greater variance as well. A similar but less sudden trend can be noted in Figure \ref{fig:plot2_cars3d_irs} on Cars3D: increasing $\beta$ leads to increased score, but the median score remains approximately steady when increasing $\gamma$. In general, therefore, it seems that \arnold is less sensitive to $\gamma$ than $\beta$-VAE is to $\beta$ when it comes to the considered datasets and disentanglement metrics. \\

\begin{figure}[t]
	\centering
	\includegraphics[page=1,width=0.8\textwidth]{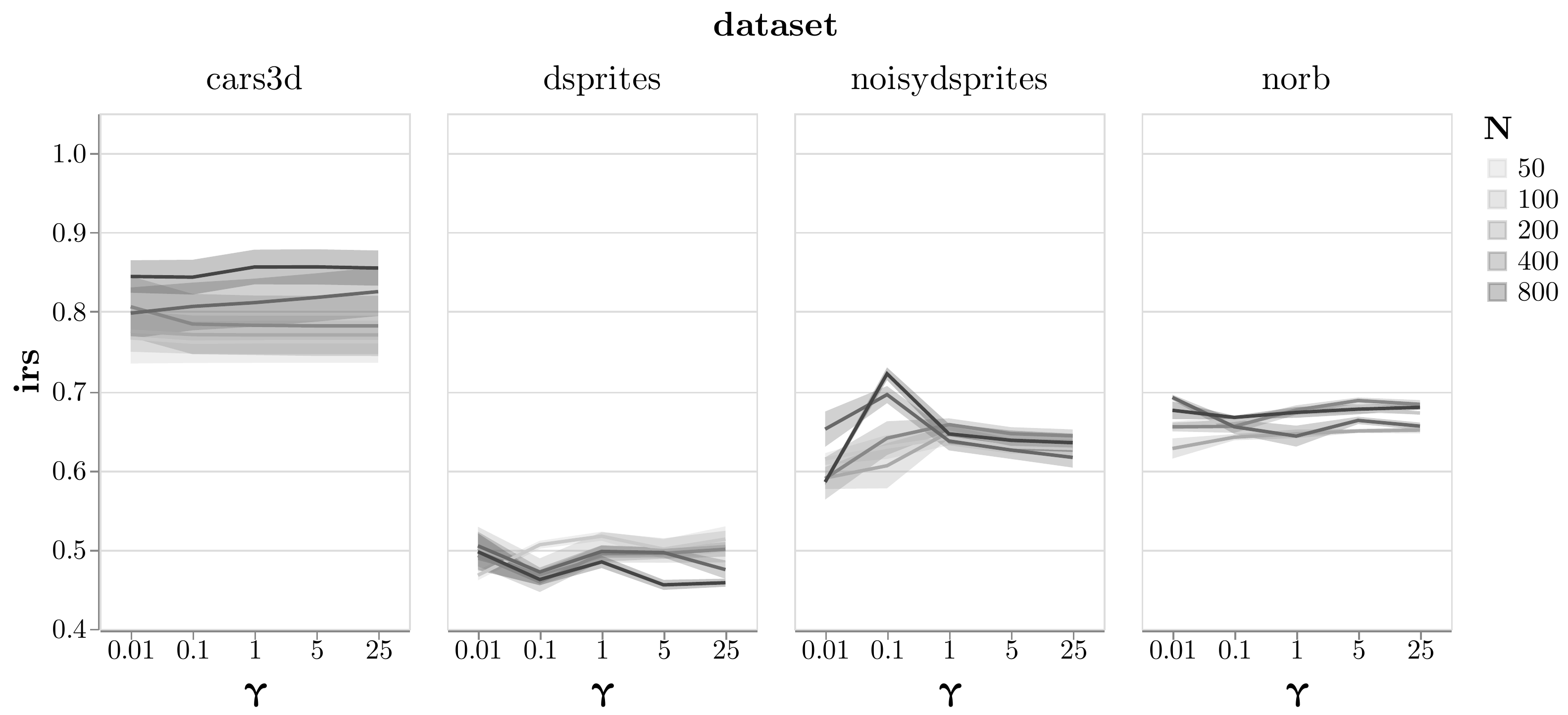}
	\caption{Line chart of the mean IRS score of a 2-layer \arnold architecture, with $s_1=10$ and $s_2=5$, according to the hyperparameter $\gamma$ for each dataset and for each number $N$ of training points. The error bands is set to the value of standard error, extending from the mean. The variance is due to five different random seeds. The higher the curve, the better.}
	\label{fig:plot7_irs}
\end{figure}

The influence of $\gamma$ is also studied separately for each $N \in \{50,100,200,400,800\}$. Figure \ref{fig:plot7_irs} plots the IRS score attained by \arnold against the hyperparameter $\gamma$ according to the number $N$ of training data points. The variance is due to five random seeds. It can be noted from the plot that most lines are roughly horizontal, meaning that varying $\gamma$ does not significantly affect the IRS score and that this behavior is shared by all considered number of training points. These results corroborate the findings of the previous set of experiments: \arnold's disentangling performance tends to remain steady as its $\gamma$ hyperparameter varies, contrary to the behavior of $\beta$-VAE with respect to its hyperparameter $\beta$, which greatly influences its performance. \\

Finally, we compare random initialization to layer-wise kernel PCA initialization. Table \ref{tab:init} shows the mean disentanglement scores attained on Cars3D over 5 random seeds. 
In this experiment, initializing the hidden units using kernel PCA in a layer-wise manner outperforms $\beta$-VAE and it provides the additional benefit of increased reliability as standard deviation is zero because no random seed is employed. 
In particular, layer-wise kernel PCA initialization on average achieves slightly lower IRS score compared to random initialization and it significantly outperforms both random initialization and $\beta$-VAE on the MIG metric. 
Overall, when it comes to the variance of the results, \arnold with layer-wise kernel PCA initialization compares favorably to $\beta$-VAE, successfully addressing the core issue of reliability of VAE-based methods brought up by \citep{locatello}.

\begin{table}[t]
	\centering
	\begin{tabular}{llll}
		\toprule
		{} &               Type &            IRS &            MIG \\
		\midrule
		&           Random initialization &  0.843 $\pm$ 0.044 &  0.012 $\pm$ 0.010 \\
		&        Layer-wise kernel PCA initialization &  0.785 $\pm$ 0.000 &  0.040 $\pm$ 0.000 \\
		&                     $\beta$-VAE &  0.752 $\pm$ 0.145 &  0.010 $\pm$ 0.011 \\
		\bottomrule
	\end{tabular}
	\caption{Comparison of disentangling performance of two \arnold models with different initialization and $\beta$-VAE on the Cars3D dataset with $N=800$. Mean IRS and MIG scores are reported with their standard deviation, which, in this case, is only due to 5 different random seeds. The employed \arnold model has two RBF layers ($\sigma^2=50$) with $s_1=10$ and $s_2=5$. The $\beta$-VAE model has $\beta=4$. Higher is better on all metrics.}
	\label{tab:init}
\end{table}
\section{Conclusion}
\label{sec:conclusion}

In this work we have proposed to reformulate the deep restricted kernel machine framework for kernel PCA \citep{drkm} into a constrained optimization problem with orthogonality constraints on the latent variables. At the same time, we have described a training algorithm that learns the hidden features in an end-to-end manner instead of layer-wise by employing a quadratic penalty optimization algorithm with warm start. We have then showed how the proposed method can be applied to denoising and to the problem of learning disentangled features in an unsupervised manner without any prior knowledge on the generative factors. In the former task, we studied the role of each principal component in every layer showing that components in the first layer perform lower-level feature detection, while components in the second layer employ the representation learned by lower layers and extract more global features, more accurately reproducing the original data distribution. In our experiments in the task of disentangled factor learning, the proposed \arnold method quantitatively performed similarly overall compared to $\beta$-VAE \citep{betavae} on four benchmark datasets with respect to a number of different disentanglement metrics when few training points are available. In addition, regarding the issue raised in \citep{locatello} that performance of state-of-the-art approaches to disentangled factor learning based on VAEs greatly varies when changing random seed or hyperparameter, \arnold was less sensitive to randomness and hyperparameter choice compared to $\beta$-VAE. In particular, the variance due to \arnold's hyperparameter $\gamma$ was smaller than the variance due to the hyperparameter $\beta$ in $\beta$-VAE and it was shown that \arnold with deterministic layer-wise kernel PCA initialization attained favorable scores without the need of a random seed, considerably improving the reproducibility of the results. Finally, the experimental analysis of the number of layers of \arnold indicates that adding a layer can increase the disentangling performance, as it was observed that a 2-layer architecture is a better choice than a single layer one. Nevertheless, 3-layer models did not consistently perform better than 2-layer models. This result does not rule out the influence of other factors, as, for example, more challenging datasets may benefit from additional layers. In future work, applying \arnold to more complicated datasets may be useful to better understand the role of the number of layers. Furthermore, it would also be interesting to investigate more advanced constrained optimization algorithms that could be useful to boost training efficiency.

\section*{Acknowledgments}
EU: The research leading to these results has received funding from
the European Research Council under the European Union's Horizon
2020 research and innovation program / ERC Advanced Grant E-DUALITY
(787960). This paper reflects only the authors' views and the Union
is not liable for any use that may be made of the contained information.
Research Council KUL: Optimization frameworks for deep kernel machines C14/18/068.
Flemish Government:
FWO: projects: GOA4917N (Deep Restricted Kernel Machines:
Methods and Foundations), PhD/Postdoc grant
Impulsfonds AI: VR 2019 2203 DOC.0318/1QUATER Kenniscentrum Data
en Maatschappij.
This research received funding from the Flemish Government (AI Research Program). Johan Suykens and Panagiotis Patrinos are affiliated to Leuven.AI - KU Leuven institute for AI, B-3000, Leuven, Belgium.
Ford KU Leuven Research Alliance Project KUL0076 (Stability analysis
and performance improvement of deep reinforcement learning algorithms).
EU H2020 ICT-48 Network TAILOR (Foundations of Trustworthy AI - Integrating Reasoning, Learning and Optimization).
This work was supported by the Research Foundation Flanders (FWO) research projects G086518N, G086318N, and G0A0920N; Fonds de la Recherche Scientifique — FNRS and the Fonds Wetenschappelijk Onderzoek — Vlaanderen under EOS project no 30468160 (SeLMA).
The computational infrastructure and services used in this work were provided by the VSC (Flemish Supercomputer Center), funded by the Research Foundation - Flanders (FWO) and the Flemish Government.

\bibliography{references}

\clearpage

\appendix
\appendixpage
\section{Further details on the experimental evaluation}
\label{sec:app}

\renewcommand{\thefigure}{A.\arabic{figure}}
\setcounter{figure}{0}
\renewcommand{\thetable}{A.\arabic{table}}
\setcounter{table}{0}
\renewcommand{\theequation}{A.\arabic{equation}}
\setcounter{equation}{0}

In Section \ref{sec:exp} we outlined the most representative experimental results and discussed them. This section goes into detail about the setup of the experiments and includes all our experimental results. In particular, it contains plots that show \arnold's performance across disentanglement metrics, dataset and hyperparameters. Plots comparing \arnold's performance to $\beta$-VAE's performance are also given.

\subsection{Datasets}
\label{app:datasets}
Four datasets are used  in the experimental evaluation: the Cars3D dataset \citep{cars3d}, the dSprites dataset \citep{betavae}, the SmallNORB dataset \citep{lecun2004} and a noisy version of dSprites introduced in \citep{locatello} obtained by replacing the background pixels with Gaussian noise with zero mean and unit variance. Each data point is generated deterministically from a tuple of factors of variations. The number of the factors of variations varies across datasets. Each factor of variation can take a finite number of values, so the number of training points is fixed and is the number of all possible combinations of the factors of variations. Table \ref{tab:datasets} summarizes the key properties of the datasets considered in the experimental evaluation.

\subsection{Experiments hyperparameters}
The chosen algorithm for the unconstrained optimization problems is Adam \citep{adam} with learning rate fixed to $10^{-3}$ and in Algorithm \ref{alg:train} we set $\mu_0=1, \tau_0=0.1$ and $p=8$. The maximum number of outer iterations was set to 2 for $N=50$ and $N=100$, to 4 for $N=200$ and to 7 for $N=400$ and $N=800$. The maximum number of inner iterations was set to 500.\\

The $\beta$-VAE encoder is constructed following the architecture proposed in \citep{betavae}. The inputs are images $x$ of dimension $c \times 64 \times 64$, where $c$ is 1 for dSprites and SmallNORB and is 3 for Cars3D and noisy dSprites. We encode $x$ using the following network: \texttt{conv} 32 $\rightarrow$ \texttt{conv} 32 $\rightarrow$ \texttt{conv} 64 $\rightarrow$ \texttt{conv} 64 $\rightarrow$ \texttt{conv} 256 $\rightarrow$ \texttt{FC} 256 $\times$ 20, where each \texttt{conv} block is a $4\times4$ convolution with stride 2 except the last block with stride 4. Each \texttt{conv} block is followed by \texttt{ReLU}. The decoder is the deconvolution reverse of the encoder.

\begin{table}[h!]
	\begin{tabular}{l|l|l|l|l}
		Dataset        & Input dimensions & \begin{tabular}[c]{@{}l@{}}\# factors of\\ variation\end{tabular} & \begin{tabular}[c]{@{}l@{}}Meaning of the factors of variation\\ and \# possible values\end{tabular}                                                                                                          & \begin{tabular}[c]{@{}l@{}}Total \# \\ data points\end{tabular} \\ \hline
		Cars3D         & 3 x 64 x 64      & 3                                                                 & \begin{tabular}[c]{@{}l@{}}- elevation (4 possible values)\\ - azimuth (24 possible values)\\ - object type (183 possible values)\end{tabular}                                                                & 17568                                                           \\ \hline
		dSprites       & 1 x 64 x 64      & 5                                                                 & \begin{tabular}[c]{@{}l@{}}- shape (3 possible values)\\ - scale (6 possible values)\\ - orientation (40 possible values)\\ - position x (32 possible values)\\ - position y (32 possible values)\end{tabular} & 737280                                                          \\ \hline
		Noisy dSprites & 3 x 64 x 64      & 3                                                                 & \begin{tabular}[c]{@{}l@{}}- shape (3 possible values)\\ - scale (6 possible values)\\ - orientation (40 possible values)\\ - position x (32 possible values)\\ - position y (32 possible values)\end{tabular} & 737280                                                          \\ \hline
		SmallNORB      & 1 x 64 x 64      & 4                                                                 & \begin{tabular}[c]{@{}l@{}}- category (5 possible values)\\ - elevation (9 possible values)\\ - azimuth (18 possible values)\\ - lighting condition (6 possible values)\end{tabular}                          & 4860                                                           
	\end{tabular}
	\caption{Key properties of the datasets used in the experimental evaluation.}
	\label{tab:datasets}
\end{table}

\clearpage

\FloatBarrier
\subsection{Additional plots of the experimental results}
\label{app:plots}

\FloatBarrier
\subsubsection{Investigation on the role of the number of selected principal components}
\null
\vfill
\begin{figure}[!htb]
	\centering

	\begin{subfigure}[b]{0.8\textwidth}
		\includegraphics[page=1,width=\textwidth]{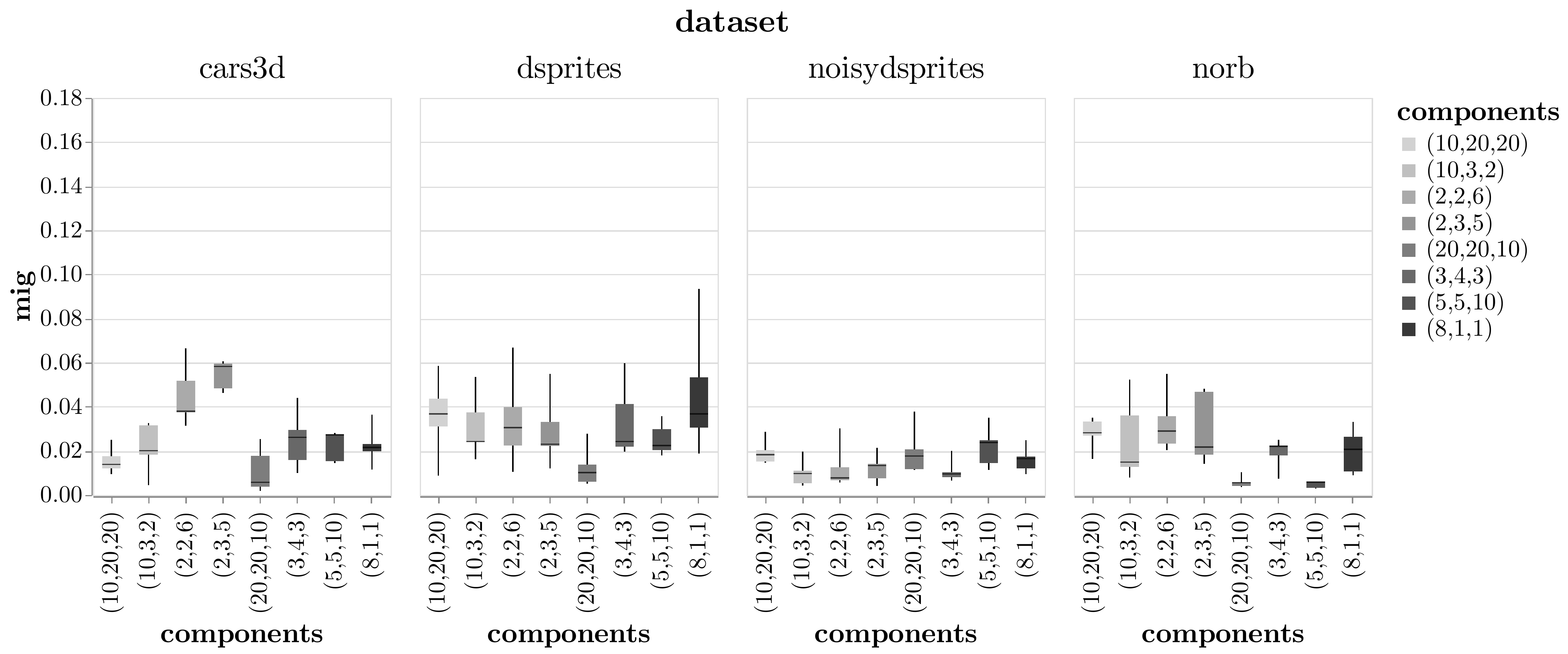}
		\caption{metric = MIG}
		\label{fig:plot8_mig}
	\end{subfigure}
	
	\begin{subfigure}[b]{0.8\textwidth}
		\includegraphics[page=1,width=\textwidth]{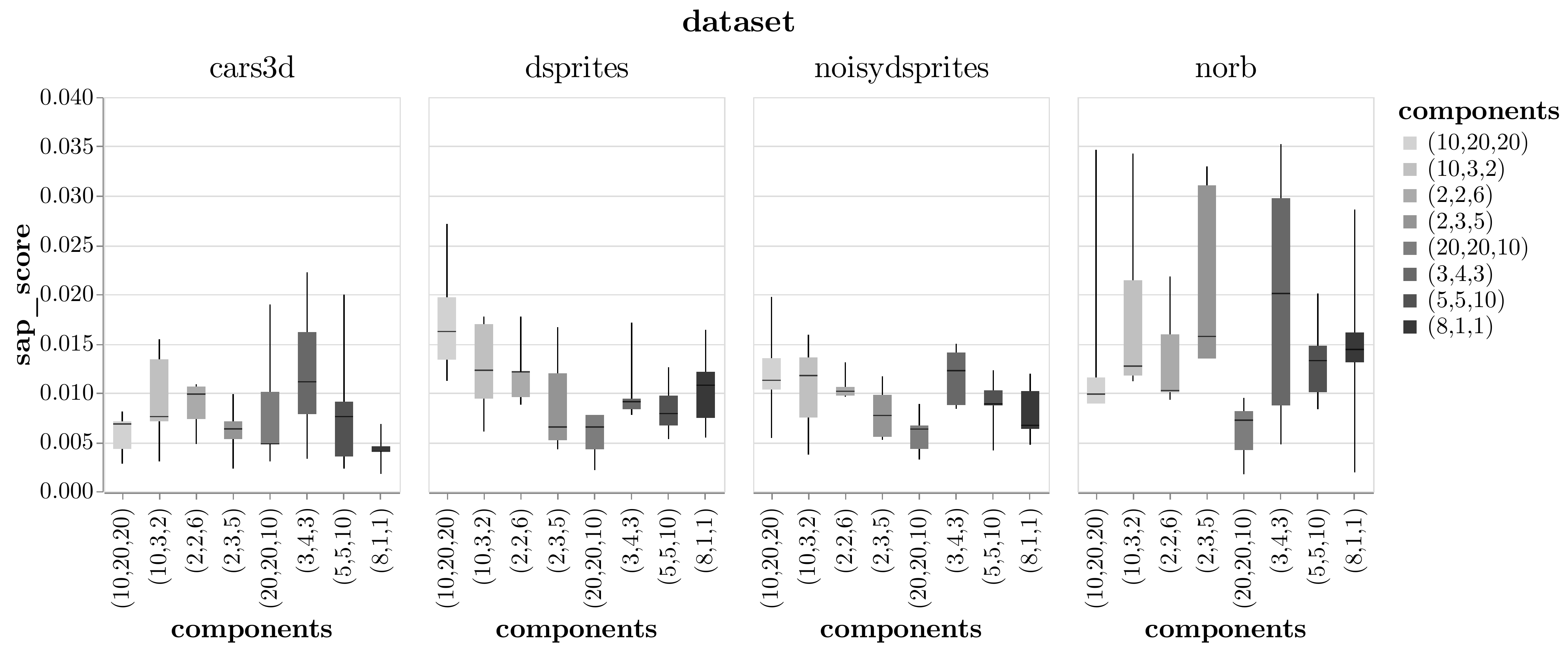}
		\caption{metric = SAP}
		\label{fig:plot8_sap_score}
	\end{subfigure}
	
	\caption{Boxplot of the disentanglement score of a 3-layer \arnold architecture according to the number of selected principal components for each dataset. The tuples in the labels are of the form $(s_1,s_2,s_3)$. Lower and upper box boundaries first and third quartile, respectively, line inside box median, lower and upper error lines minimum and maximum value, respectively. The results are shown over five random seeds. The plot for the IRS metric is in Figure \ref{fig:plot8_irs}.}
	\label{fig:plot8app}
\end{figure}
\vfill

\clearpage
\FloatBarrier
\subsubsection{Investigation on the role of the number of layers}
\null
\vfill
\begin{figure}[!htb]
	\centering
	\includegraphics[page=1,width=0.9\textwidth]{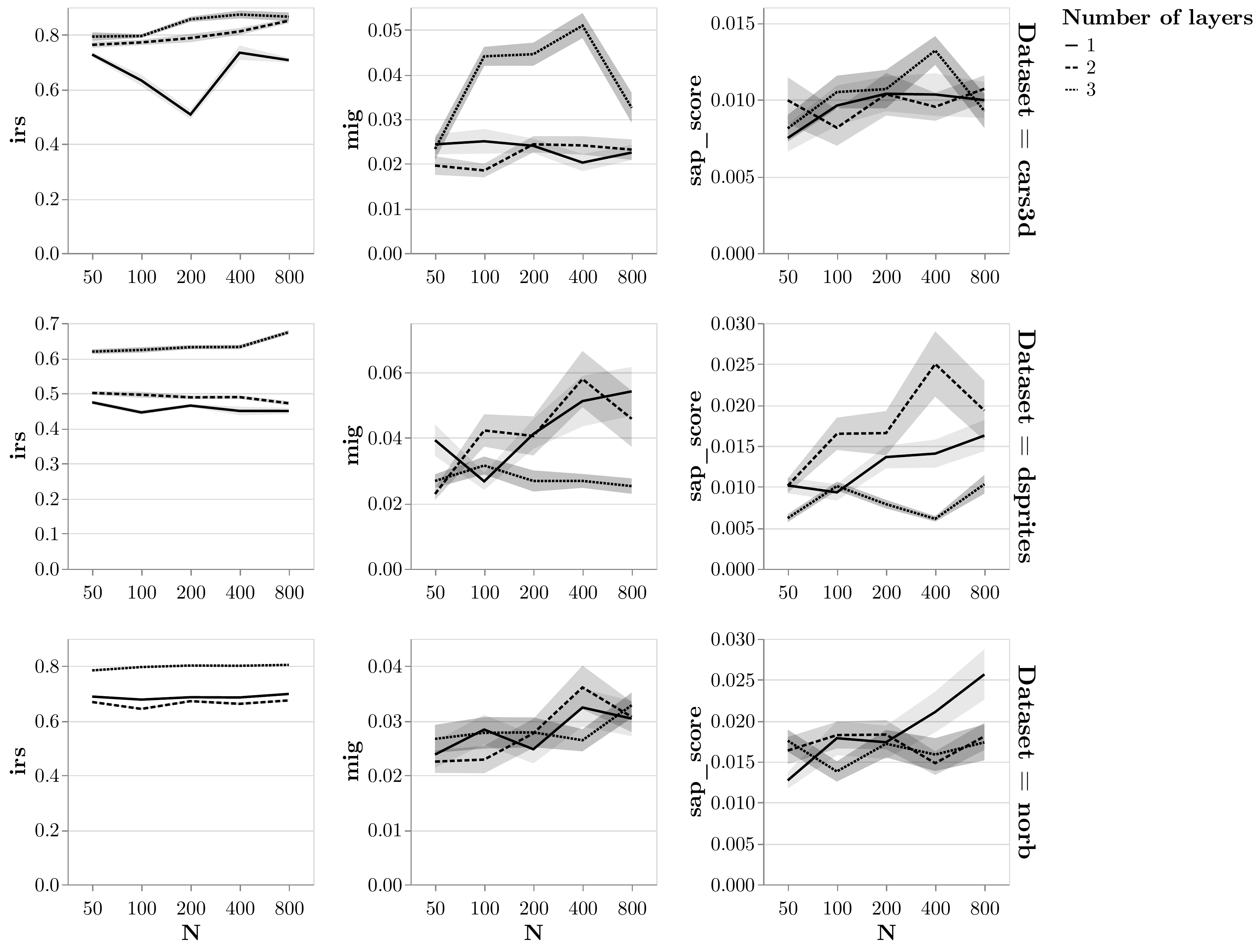}
	\caption{Line chart of the mean disentanglement score of different \arnold architectures according to the number $N$ of training data points and the number $n_\text{layers}$ of layers for all datasets. For the 1-layer architecture, $s_1=10$, for the 2-layer one, $s_1=10$ and $s_2=5$ and for the 3-layer one, $s_1=2$, $s_2=2$ and $s_3=6$. The size of each error band is set to the value of standard error, extending from the mean. The variance is due to five different random seeds and different $\gamma$.}
	\label{fig:plot10}
\end{figure}
\vfill

\clearpage
\FloatBarrier
\subsubsection{Comparison of \arnold and $\beta$-VAE according to the number of training points}
\null
\vfill
\begin{figure}[h]
	\centering
	\includegraphics[page=1,width=0.7\textwidth]{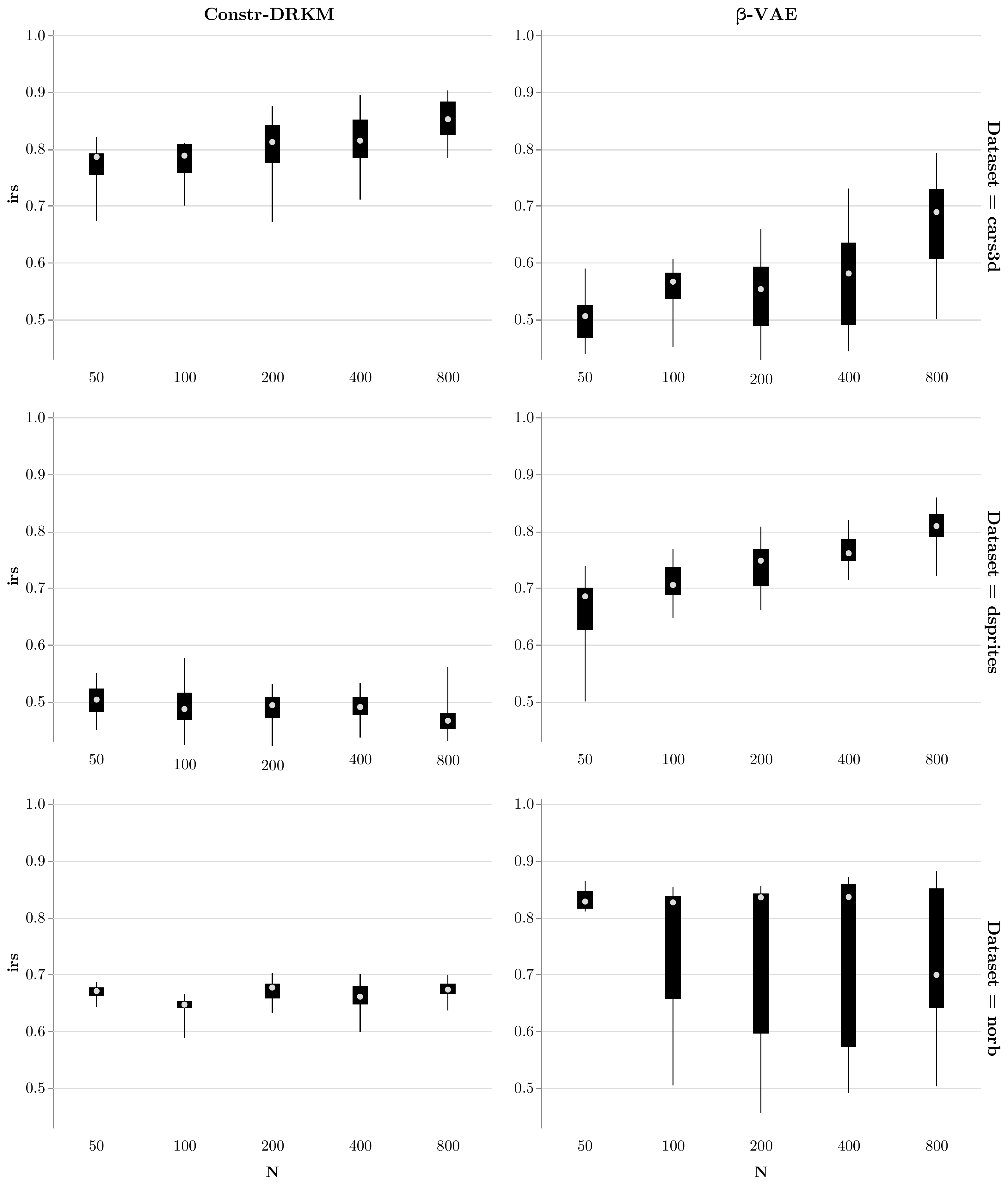}
	\caption{Boxplots of the IRS score of a 2-layer \arnold, with $s_1=10$ and $s_2=5$, and of a $\beta$-VAE model according to the number of training points for each dataset. Disentanglement scores are shown across all choices of the hyperparameter $\gamma$ for \arnold and of $\beta$ for $\beta$-VAE and across five random seeds. The boxes have the same structure as in Figure \ref{fig:plot1}.}
	\label{fig:plot3_irs}
\end{figure}
\vfill

\begin{figure}[!htb]
	\centering
	\includegraphics[page=1,width=0.7\textwidth]{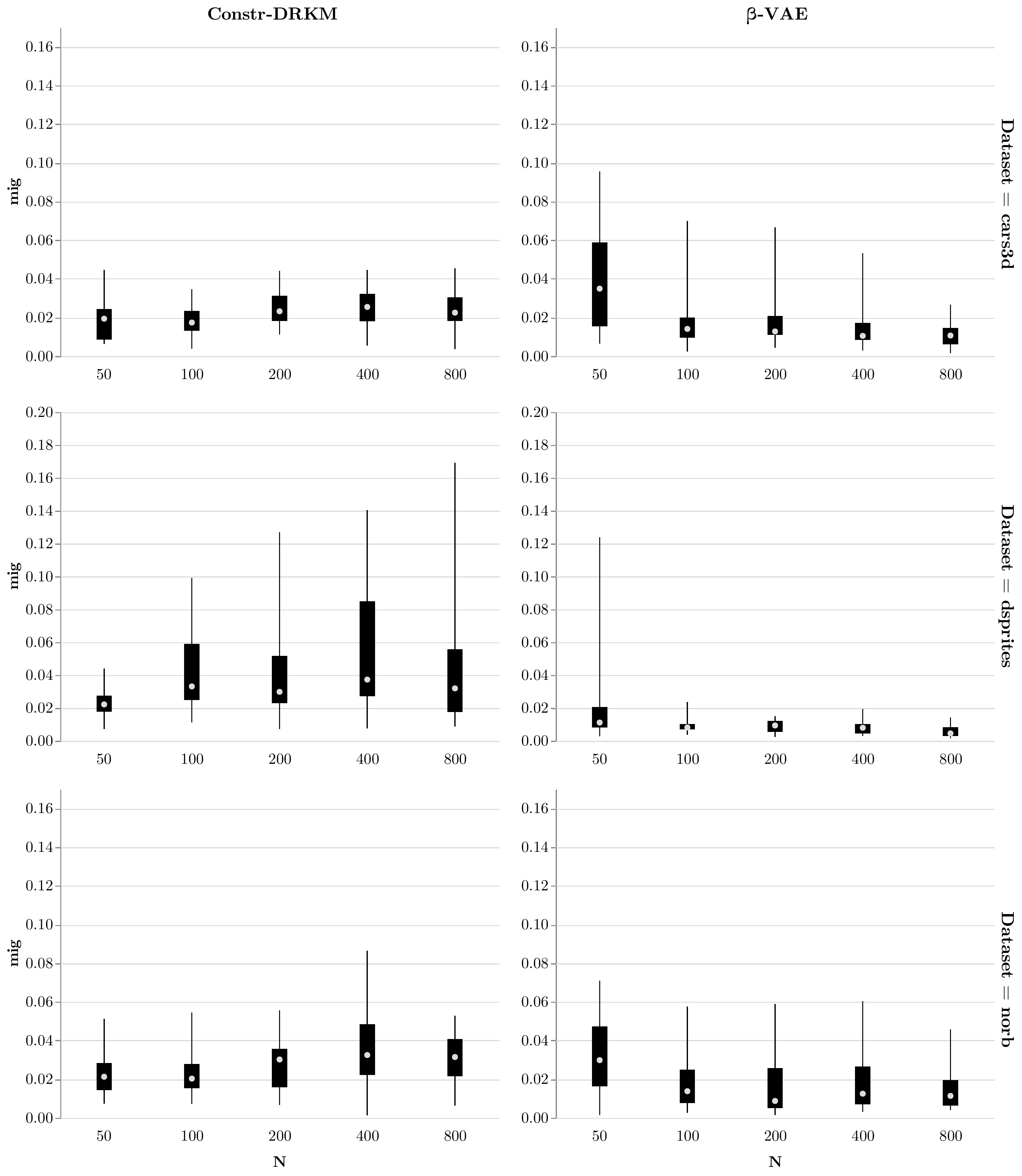}
	\caption{Boxplots of the MIG score of a 2-layer \arnold, with $s_1=10$ and $s_2=5$, and of a $\beta$-VAE model according to the number of training points for each dataset. Disentanglement scores are shown across all choices of the hyperparameter $\gamma$ for \arnold and of $\beta$ for $\beta$-VAE and across five random seeds. The boxes have the same structure as in Figure \ref{fig:plot1}.}
	\label{fig:plot3_mig}
\end{figure}

\begin{figure}[!htb]
	\centering
	\includegraphics[page=1,width=0.7\textwidth]{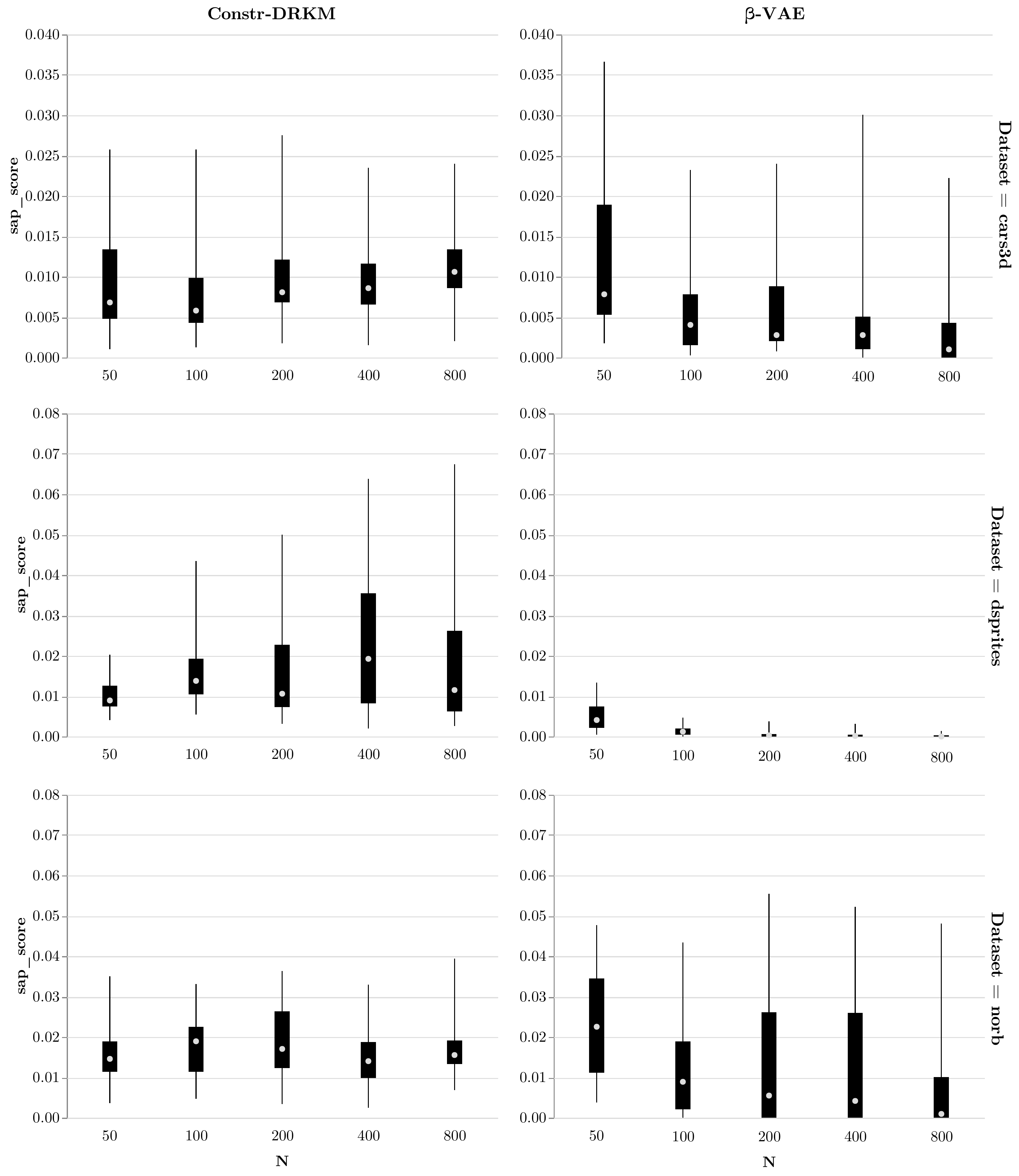}
	\caption{Boxplots of the SAP score of a 2-layer \arnold, with $s_1=10$ and $s_2=5$, and of a $\beta$-VAE model according to the number of training points for each dataset. Disentanglement scores are shown across all choices of the hyperparameter $\gamma$ for \arnold and of $\beta$ for $\beta$-VAE and across five random seeds. The boxes have the same structure as in Figure \ref{fig:plot1}.}
	\label{fig:plot3_sap_score}
\end{figure}

\FloatBarrier
\subsubsection{Comparison of \arnold and $\beta$-VAE according to the hyperparameter $\gamma$ and $\beta$, respectively}
\null
\vfill
\begin{figure}[!htb]
	\centering
	\includegraphics[page=1,width=0.7\textwidth]{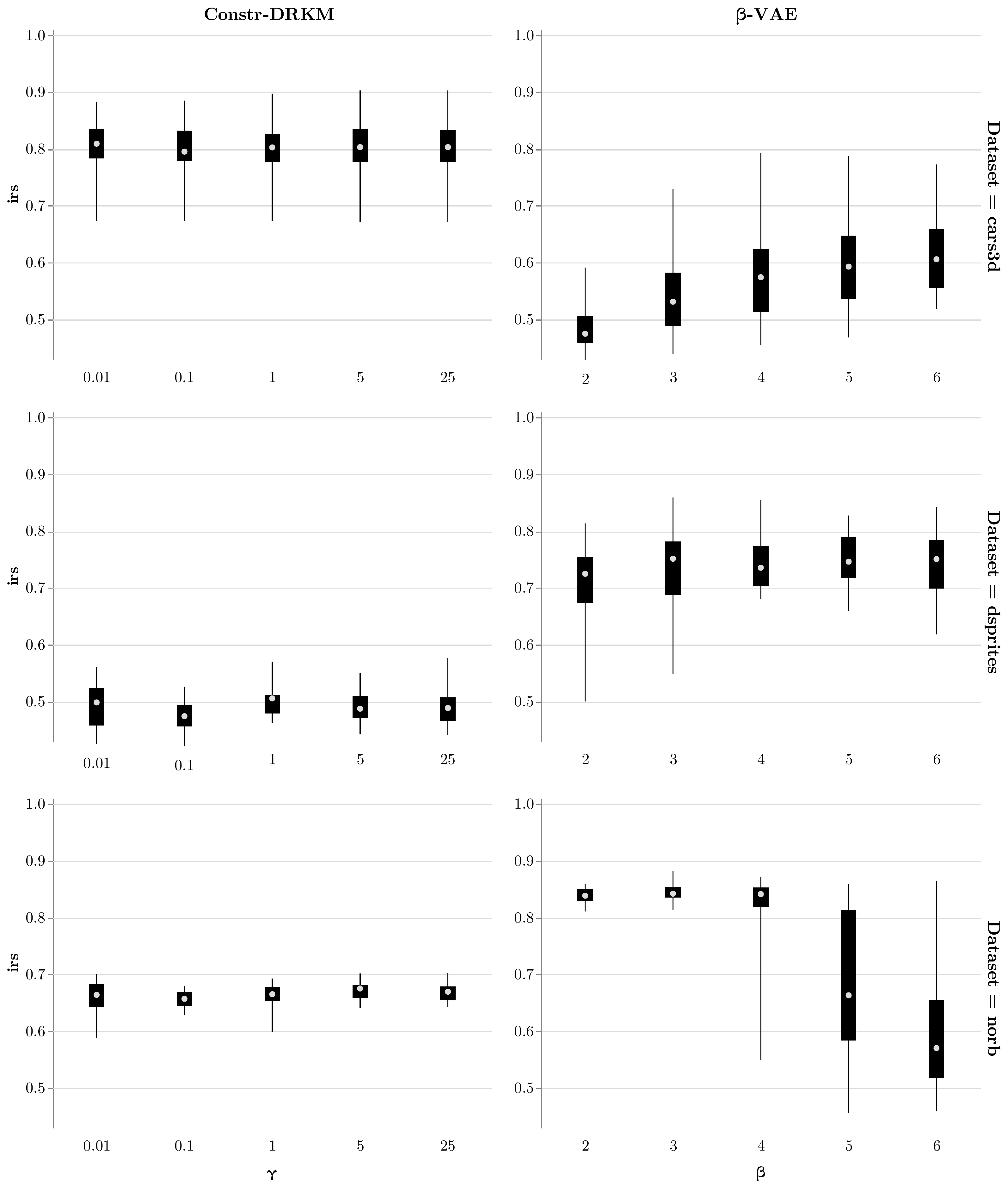}
	\caption{Boxplots of the IRS score of a 2-layer \arnold, with $s_1=10$ and $s_2=5$, and of a $\beta$-VAE model according to the hyperparameter $\gamma$ for \arnold and of $\beta$ for $\beta$-VAE on the cars3D and SmallNORB datasets. Disentanglement scores are shown across all number $N$ of training points and across five random seeds. The boxes have the same structure as in Figure \ref{fig:plot1}.}
	\label{fig:plot5_irs}
\end{figure}
\vfill

\begin{figure}[!htb]
	\centering
	\includegraphics[page=1,width=0.7\textwidth]{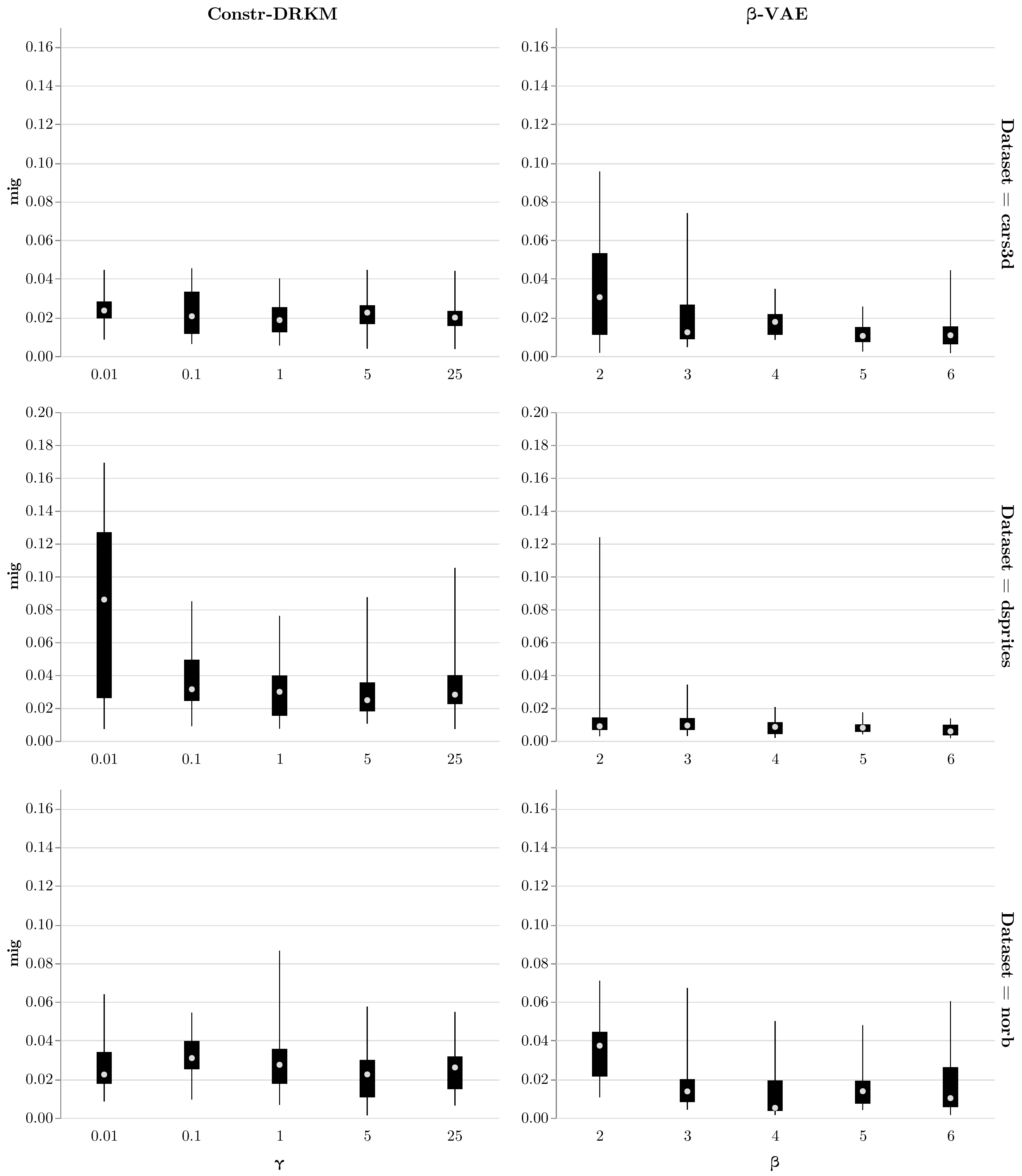}
	\caption{Boxplots of the MIG score of a 2-layer \arnold, with $s_1=10$ and $s_2=5$, and of a $\beta$-VAE model according to the hyperparameter $\gamma$ for \arnold and of $\beta$ for $\beta$-VAE on the cars3D and SmallNORB datasets. Disentanglement scores are shown across all number $N$ of training points and across five random seeds. The boxes have the same structure as in Figure \ref{fig:plot1}.}
	\label{fig:plot5_mig}
\end{figure}

\begin{figure}[!htb]
	\centering
	\includegraphics[page=1,width=0.7\textwidth]{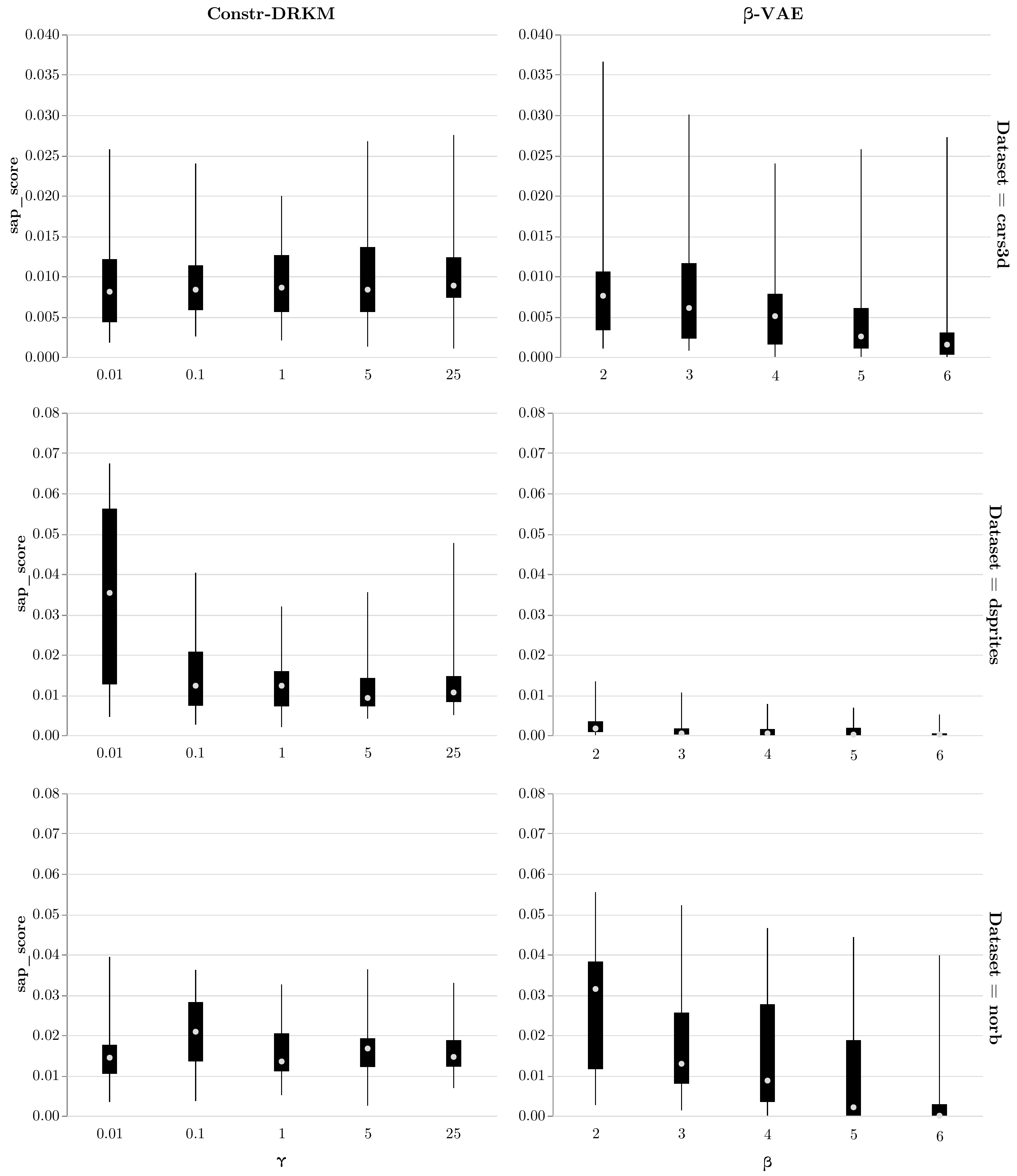}
	\caption{Boxplots of the SAP score of a 2-layer \arnold, with $s_1=10$ and $s_2=5$, and of a $\beta$-VAE model according to the hyperparameter $\gamma$ for \arnold and of $\beta$ for $\beta$-VAE on the cars3D and SmallNORB datasets. Disentanglement scores are shown across all number $N$ of training points and across five random seeds. The boxes have the same structure as in Figure \ref{fig:plot1}.}
	\label{fig:plot5_sap_score}
\end{figure}

\FloatBarrier
\subsubsection{Investigation on the influence of the hyperparameter $\gamma$}
\null
\vfill
\begin{figure}[!htb]
	\centering
	\includegraphics[page=1,width=0.9\textwidth]{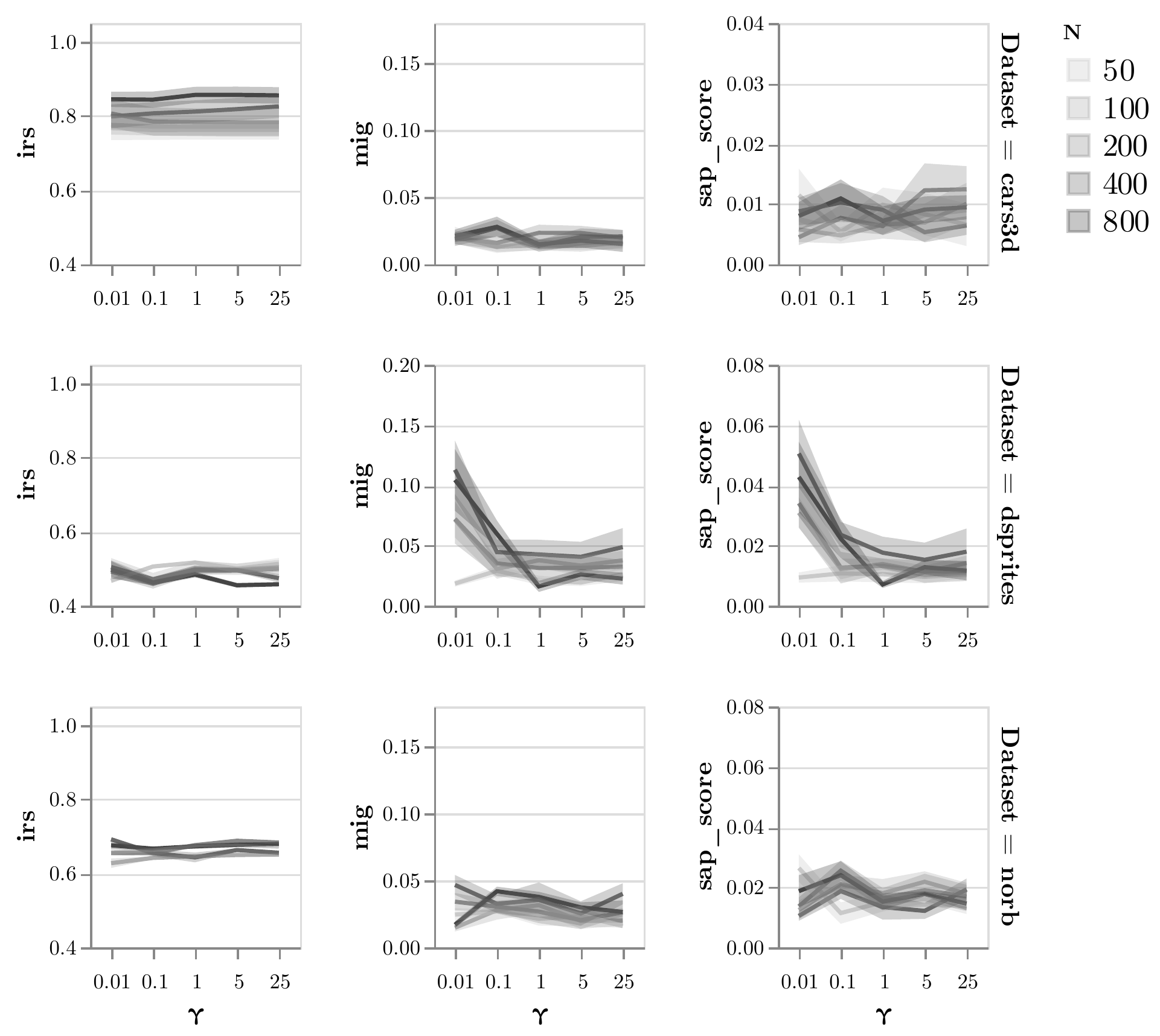}
	\caption{Line chart of the mean disentanglement score of a 2-layer \arnold architecture, with $s_1=10$ and $s_2=5$, according to the hyperparameter $\gamma$ for each dataset and for each number $N$ of training points. The size of each error band is set to the value of standard error, extending from the mean. The variance is due to five different random seeds.}
	\label{fig:plot6}
\end{figure}
\vfill

\end{document}